\documentclass{article}
\newcommand{\dbname}{CSTS{}} 
\newcommand{\R}{\mathbb{R}}

\newcommand{\nVariates}{V}
\newcommand{\eigenvectorMatrix}{U}
\newcommand{\eigenvaluesMatrix}{\Lambda}
\newcommand{\eigenvalue}[1][i]{\lambda_{#1}}
\newcommand{\corrTransformationMatrix}{W}
\newcommand{\nSegments}{M}
\newcommand{\segIndex}{m}

\newcommand{\nClusters}{K}
\newcommand{\clusterIndex}{k}

\newcommand{\segment}[1][\segIndex]{\mathbf{S}_{#1}}

\newcommand{\correlationMatrix}[1][\segIndex]{\mathbf{A}_{#1}}
\newcommand{\correlationMatrixElement}{a}

\newcommand{\segmentation}[1][]{\eta_{#1}}
\newcommand{\partitionIndex}{i}

\newcommand{\partition}[1][\partitionIndex]{\mathbf{\pi_{#1}}}

\newcommand{\clusterIndices}[1][\clusterIndex]{\phi_{#1}}

\newcommand{\oldCluster}[1][]{\text{cluster check}}

\newcommand{\patternIndex}{\ell}
\newcommand{\canonicalPattern}[1][\patternIndex]{\mathbf{P}_{#1}}
\newcommand{\canonicalPatternElement}{p}

\newcommand{\relaxedPattern}[1][\patternIndex]{\mathbf{P}'_{#1}}
\newcommand{\toleranceBands}{\mathcal{B}}

\newcommand{\internalIndex}[1][i]{I_{#1}}

\newcommand{\distanceMeasure}[1][]{d_{#1}(\correlationMatrix, \relaxedPattern)}

\newcommand{\silhouetteIndex}[1][\segIndex]{sil_{#1}}
\newcommand{\intraClusterDistance}[1][\segIndex]{a_{#1}}
\newcommand{\interClusterDistance}[1][\segIndex]{b_{#1}}

\newcommand{\clusterDistance}[1][\corrMatY]{d(\correlationMatrix, #1)}
\newcommand{\averageClusterDistance}[1][\clusterIndex]{\sigma_{#1}}
\newcommand{\clusterCentroid}[1][\clusterIndex]{\correlationMatrix[C_{#1}]}

\usepackage{arxiv}

\usepackage[utf8]{inputenc} 
\usepackage[T1]{fontenc}    
\usepackage{hyperref}       
\usepackage{url}            
\usepackage{booktabs}       
\usepackage{amsfonts}       
\usepackage{nicefrac}       
\usepackage{microtype}      
\usepackage[dvipsnames]{xcolor}         

\usepackage{amsmath}
\usepackage[]{amsthm}
\usepackage[]{amssymb}
\usepackage{dsfont}
\usepackage{graphicx}
\usepackage{subcaption}
\usepackage{wrapfig}
\hypersetup{
    colorlinks=true,
    linkcolor=RoyalBlue,
    filecolor=magenta,
    urlcolor=RoyalBlue,
    citecolor=RoyalBlue
}
\usepackage{multirow}

\usepackage{listings}

\definecolor{CodeBackground}{rgb}{249, 250, 251}
\definecolor{CodeComment}{named}{ForestGreen}
\definecolor{CodeKeyword}{named}{RedViolet}
\definecolor{CodeString}{named}{Green}
\lstdefinestyle{pythonstyle}{
    backgroundcolor=\color{CodeBackground},   
    commentstyle=\color{CodeComment},
    keywordstyle=\color{CodeKeyword},
    numberstyle=\tiny\color{Gray},
    stringstyle=\color{CodeString},
    basicstyle=\small\ttfamily,
    breakatwhitespace=false,         
    breaklines=true,                 
    captionpos=b,                    
    keepspaces=true,                 
    numbers=left,                    
    numbersep=5pt,                  
    showspaces=false,                
    showstringspaces=false,
    showtabs=false,                  
    tabsize=2,
    language=Python
}
\title{\dbname{}: A Benchmark for the Discovery of Correlation Structures in Time Series Clustering}

\author[1]{Isabella Degen}
\author[2]{Zahraa S Abdallah}
\author[3]{Henry W J Reeve}
\author[4]{Kate Robson Brown}
\affil[1]{School of Computer Science, University of Bristol}
\affil[2]{School of Engineering Mathematics and Technology, University of Bristol}
\affil[3]{School of Artificial Intelligence, University of Nanjing}
\affil[4]{College of Engineering and Architecture, University College Dublin}

\begin{document}
\maketitle
\begin{abstract}
Time series clustering promises to uncover hidden structural patterns in data with applications across healthcare, finance, industrial systems, and other critical domains. However, without validated ground truth information, researchers cannot objectively assess clustering quality or determine whether poor results stem from absent structures in the data, algorithmic limitations, or inappropriate validation methods, raising the question whether clustering is "more art than science" (Guyon et al., 2009). To address these challenges, we introduce \dbname{} (\textbf{C}orrelation \textbf{S}tructures in \textbf{T}ime \textbf{S}eries), a synthetic benchmark for evaluating the discovery of correlation structures in multivariate time series data. \dbname{} provides a clean benchmark that enables researchers to isolate and identify specific causes of clustering failures by differentiating between correlation structure deterioration and limitations of clustering algorithms and validation methods.
Our contributions are: (1) a comprehensive benchmark for correlation structure discovery with distinct correlation structures, systematically varied data conditions, established performance thresholds, and recommended evaluation protocols; (2) empirical validation of correlation structure preservation showing moderate distortion from downsampling and minimal effects from distribution shifts and sparsification; and (3) an extensible data generation framework enabling structure-first clustering evaluation. A case study demonstrates \dbname{}'s practical utility by identifying an algorithm's previously undocumented sensitivity to non-normal distributions, illustrating how the benchmark enables precise diagnosis of methodological limitations. \dbname{} advances rigorous evaluation standards for correlation-based time series clustering.
\end{abstract}

\section{Introduction}
Clustering is an established and essential data mining approach in countless disciplines, encompassing numerous algorithms and validation techniques \cite{Fahad2014, Ezugwu2022, Xu2015, Gao2023, Jaeger2023, Paparrizos2024}. Multivariate time series clustering by correlation structures groups data based on relationship patterns between variates. This technique is critical for detecting state transitions, identifying anomalies, and understanding complex temporal dynamics across domains including biology \cite{Chandereng2020}, finance \cite{Marti2021}, and industrial systems \cite{Iglesias2013}. Clustering, an unsupervised machine learning technique, autonomously discovers intrinsic patterns without requiring labelled training data. This makes it valuable for exploratory analysis and novel discovery. The reliability of these discoveries is critical in high-stakes environments, where misidentifying state transitions can have serious consequences.

Despite its importance, validating that clustering algorithms correctly identify groupings based on structural data properties (e.g. topological, geometrical, statistical) rather than arbitrary patterns remains challenging, as it requires hard-to-come-by benchmarking datasets with ground truth labels \cite{Xu2015, Thrun2021, Herrmann2024}. Without ground truth information related to structural properties, researchers cannot determine whether a failure to discover specific structures in the data stems from their absence, algorithmic limitations, or inappropriate validation metrics. This ambiguity hampers the rigorous evaluation and benchmarking of clustering methods, leading to potentially misleading conclusions about their effectiveness.

Current clustering benchmarks are based on classification datasets such as UCR \cite{Dau2019}, whose boundaries align with human understanding of classes, not necessarily structural data properties. This can cause algorithms to perform poorly not because they do not find meaningful structures, but because the expected classification labels do not match the structures that the algorithm naturally discovers \cite{Javed2020, Paparrizos2023,Thrun2021}. Despite these insights, the field has not systematically organised clustering techniques by the structural properties they are designed to detect, nor developed structure-specific benchmarks to validate them, as has been pointed out for topological structures by \cite{Herrmann2024}.

We address this critical gap by introducing \dbname{} (\textbf{C}orrelation \textbf{S}tructures in \textbf{T}ime \textbf{S}eries), a structure-first benchmark for correlation-based multivariate time series clustering. Unlike previous approaches, \dbname{} is designed around a complete set of well-defined correlation structures (a type of statistical structure) rather than arbitrary classification boundaries. \dbname{} includes ground truth clustering labels for controlled data variations (distribution shifts, sparsification, and downsampling) as well as predefined lower quality clustering labels. Our benchmark enables researchers to systematically and rigorously evaluate both clustering algorithms and validation methods under varying data conditions. \dbname{} represents to our knowledge the first correlation structure-specific evaluation framework, enabling rigorous assessment of both clustering algorithms and validation methods. 

\dbname{} makes three main contributions to the field of time series clustering:

\begin{enumerate}
    \item A comprehensive benchmark for the discovery of correlation structures in time series data with distinct correlation structures, systematically varied data conditions (distribution shifts, sparsification, downsampling), labels for ground truth and controlled degraded clustering results, established performance thresholds for correlation structures, and recommended evaluation protocols.
    \item Empirical validation of correlation structure preservation showing moderate distortion from downsampling and minimal effects from distribution shifts and sparsification; additionally showing that negative correlations are more vulnerable to distortion, Spearman correlation consistently outperforms alternatives, and reliable correlation structure estimation requires a minimum segment length.
    \item An extensible data generation framework that enables customisation of correlation structures and related parameters, facilitates structure-first clustering evaluation, and provides a template for establishing rigorous clustering benchmarks beyond correlation structures.
\end{enumerate}

The remainder of this paper is organised as follows. Section~\ref{sec:related-work} positions our work within the clustering literature. Section~\ref{sec:benchmark-design} describes the design and generation of the dataset. Section~\ref{sec:benchmark-validation} presents correlation structure validation findings. Section~\ref{sec:benchmark-usage} details the dataset usage guidelines. Section~\ref{sec:case-study} demonstrates a practical application. Sections~\ref{sec:limitations} and \ref{sec:conclusion} address limitations and conclude the paper.

\section{Related Work}\label{sec:related-work}
Clustering research has produced numerous algorithms and validation techniques, yet it remains unclear which methods work best for a specific dataset \cite{Jaeger2023, Ezugwu2022}. Researchers have demonstrated that different algorithms excel at detecting specific structural properties: k-means performs best with spherical clusters, DBSCAN excels at density-based structures, and hierarchical methods effectively capture nested relationships \cite{Xu2015, Thrun2021}. Recent work has emphasised the importance of a topological perspective on clustering \cite{Herrmann2024, Niyogi2011}, arguing that clustering fundamentally attempts to separate data into connected components. Despite these insights, the field has not systematically organised clustering techniques by the structural properties they are designed to detect, nor developed structure-specific benchmarks to validate them \cite{Herrmann2024}. This leads to customised approaches for each application context and proliferating claims of superior metrics or algorithms without clarifying which structures (topological, geometrical, statistical) they excel at (e.g. \cite{Liu2013, Huang2022, Iglesias2013, Inoue2024}). 

Time series clustering benchmarking studies \cite{Javed2020, Paparrizos2023} usually rely on classification datasets such as the UCR archive \cite{Dau2019}, introducing a critical methodological issue: classification boundaries are defined by humans and do not necessarily reflect structural properties \cite{Herrmann2024}. Algorithms may perform poorly not because they fail to discover meaningful structures, but because the expected classifications might not match such structures \cite{Javed2020, Paparrizos2023,Thrun2021}. Consequently, whichever algorithm detects the most represented structure in the UCR datasets is evaluated as "best" regardless of its applicability to specific structures. Such fundamental issues have led researchers to debate whether clustering is a science or an art \cite{Guyon2009} and characterise it as a "quagmire" \cite{Jaeger2023}, lagging behind benchmarking standards in other machine learning fields \cite{Qian2023,Thiyagalingam2022,Middlehurst2024}.

Correlation structures describe changing relationships between time series and fall into statistical data structures. Clustering by changes in correlation structures finds applications across finance \cite{Marti2021}, industrial systems \cite{Iglesias2013}, and biology \cite{Chandereng2020}. Methods such as Toeplitz Inverse Covariance-Based Clustering (TICC) \cite{Hallac2017}, network-based approaches \cite{Marti2021}, and Lag Penalized Weighted Correlation \cite{Chandereng2020} have demonstrated effectiveness in capturing such temporal relationships. However, these methods have been evaluated on domain-specific (often proprietary) datasets that have not been systematically validated for the correlation structures they include. Comparison of these methods remains challenging without benchmarks specifically designed to validate the discovery of correlation-based structures.

Clustering algorithms validation frameworks have been extensively studied for Euclidean data (geometrical structures) \cite{Vendramin2010, Arbelaitz2013, Liu2013}, but their effectiveness to validate correlation structures remains unexplored. Current validation approaches do not consider the limitations in structures that the indices are designed to validate. Recent work by \cite{Yerbury2024} demonstrated that internal validity indices are not suitable for comparing similarity paradigms, suggesting that techniques that reveal topological structures (e.g. t-SNE, UMAP) and domain knowledge are used instead. Furthermore, there is a lack of established threshold values for internal validity indices that identify a strong grouping for each structure, leading researchers to potentially interpret weak groupings as significant \cite{Kaya2022}.

Synthetic data with controlled properties has proven to be valuable for evaluating clustering algorithms \cite{Frnti2018, Thrun2020, ElAbbassi2021, Ding2024}, as it allows researchers to isolate specific factors that affect performance. However, no previous work has developed synthetic data specifically for correlation structures and has systematically examined the impact of distribution shifting, sparsification, and downsampling on these structures. This makes it impossible to establish performance thresholds and rigorously evaluate whether algorithm limitations or data characteristics are responsible for suboptimal results. 

These challenges point to a critical need for controlled, structure-specific benchmarking datasets that serve to reliably evaluate both clustering algorithms and validation methods. \dbname{} addresses this gap by providing to our knowledge the first correlation structure-specific benchmark for time series clustering, enabling rigorous evaluation of both algorithms and validation methods with verified structure preservation under controlled degradation conditions.

\section{Benchmark Design and Generation}\label{sec:benchmark-design}
\dbname{} is designed as a structure-first benchmark that enables rigorous evaluation of both clustering algorithms and validation methods by systematically controlling relevant data characteristics. Our benchmark provides key capabilities absent in existing time series benchmarks: (1) comprehensive coverage of all possible strong correlation structures for three time series variates, (2) systematic variation of data conditions, (3) predefined degraded clustering results, and (4) versions with reduced cluster, respectively, segment count. This design enables researchers to identify specific causes for suboptimal clustering and to clearly distinguish between correlation structure deterioration and algorithmic limitations. By including perfect ground truth and controlled degraded results across the Jaccard index range, \dbname{} facilitates objective comparison under standardised conditions.

\dbname{} includes two independent synthetic time series datasets (an exploratory and a confirmatory) with $30$ subjects each. We model all valid correlation structures for strong positive, negligible, and strong negative correlations (correlation coefficients $\in \{1, 0, -1\}$) for three time series variates. This results in $27$ possible patterns of which $23$ can be modelled as valid positive semi-definite correlation matrices by adjusting the correlation coefficients within the tolerance bands $\toleranceBands=\{[-1,-0.7],[-0.2,0.2],[0.7,1]\}$ that represent meaningful thresholds for strong negative, negligible and strong positive correlations.
We call these structures relaxed canonical patterns $\relaxedPattern$. Canonical due to their discrete, interpretable, and idealised nature and relaxed as we allow the coefficients to vary within $\toleranceBands$, see Appendix~\ref{app:correlation-structures}.

Each subject consists of $100$ segments of various lengths that last from $15$ minutes to $10$ hours ($900-36000$ observations at seconds sampling) using a different randomly chosen correlation pattern for each segment. We ensure that each pattern is used $4-5$ times per subject. The resulting time series consists of stationary segments with regime-switching correlation structures that represent a different event or biological state of a system. 

Our benchmark dataset comprises $12$ distinct data variants, consisting of four generation stages (raw, correlated, non-normal, downsampled), each available in three levels of data completeness (complete 100\%, partial 70\%, and sparse 10\%). The complete data variants consist of regularly sampled time series at 1-second intervals, while the partial and sparse variants represent irregularly sampled time series. Patrial variants have a mean gap between observations of $1.45$ seconds (SD $0.78$, range $1-15$), while the sparse data variants have a mean gap between observations of $10$ seconds (SD $9.48$, range $1-135$). All downsampled data variants maintain regular sampling at one-minute intervals. In raw data variants, observations are independent and identically distributed; in correlated data variants, observations become correlated between variates, losing independence; and in non-normal data variants, observations remain correlated while also losing identical distribution.

The data were generated as following: To generate the \textbf{raw} data, we randomly sampled observations for each subject from a standard normal distribution. The sampling happens segment by segment, and the segment length was picked from a configurable list of lengths (15, 20, or 30 minutes and 1, 2, 3, 4, 5, 6, 8, 10 hours) at random. Timestamps were generated using seconds as the sampling rate. The \textbf{correlated} data was created from the raw data by encoding the correlation structures into the data. For each segment, a random correlation pattern was chosen, ensuring a consistent frequency of $4-5$ for each correlation structure per subject. The correlation was achieved through: (1) calculating the eigendecomposition of each pattern's correlation matrix $\canonicalPattern = \eigenvectorMatrix \eigenvaluesMatrix \eigenvectorMatrix^T$, where $\relaxedPattern \subseteq \R^{\nVariates\times \nVariates}$ is the correlation structure to be modelled for $\nVariates$ time series variates, $\eigenvectorMatrix$ is the matrix of eigenvectors, and $\eigenvaluesMatrix$ is the diagonal matrix of eigenvalues $\eigenvalue[1],...\eigenvalue[\nVariates]$; (2) we ensure that all eigenvalues to be positive by setting any negative eigenvalues to zero: $\eigenvalue=\max(0, \eigenvalue)$; (3) constructing the correlation transformation matrix as $\corrTransformationMatrix= (\sqrt{\eigenvaluesMatrix}\odot\eigenvectorMatrix)^T$; and (4) calculating the correlated segment data as $\segment'=\segment \corrTransformationMatrix$, where $\segment$ in $\R^{t_{\segIndex}\times \nVariates}$ is a segment, and $t_{\segIndex}$ is the number of observations in this segment. The \textbf{non-normal} data was generated by shifting the distribution of each segment of the correlated data using a slight variation in the distribution parameters between subjects. The first time series variate was shifted to an extreme value distribution with shape parameters in $[-0.52, 0.07]$, shift parameters in $[0.1, 1.49]$, and scale parameters in $[0.36, 3.22]$, which make the observation values appear similar to insulin on board (IOB). The second time series variate was shifted to a negative binominal distribution with the number of successes $n=1$ and the probability p of a single success in $[0.05, 0.4]$, which makes it appear similar to carbohydrates on board (COB). Finally, the third time series variate was shifted to an extreme value distribution with shape parameters in $[0, 0.08]$, shift parameters in $[88.79, 131.99]$, and scale parameters in $[17.82, 53.53]$, making it appear similar to interstitial glucose level (IG). These parameters stem from distribution fitting real-world Type 1 Diabetes (T1D) treatment data for IOB, COB, and IG, which is our future real-world application case. Although segments encode correlation structures, the temporal dependencies of observations that are common in real-world diabetes data are not simulated. Finally, we created the seconds to minutes \textbf{downsampled} data from the non-normal data variants by aggregating the second values within each minute into a mean value. For the raw, correlated, and non-normal complete data variants, we created both partial and sparse versions from their complete counterparts. For the downsampled data variants, we created the partial and sparse versions from their corresponding non-normal data variants, as this better reflects how downsampling is typically applied. For the partial variants, we randomly dropped observations with a probability of $p=0.3$, resulting in the retention of 70\% of the observations. For the sparse data variants, the probability was $p=0.9$, resulting in the retention of 10\% of the observations. 

To facilitate the evaluation of clustering validation measures, \dbname{} simulates $66$ degraded clusterings for each subject using the following three strategies: (1) shifting the segment end index forward by a randomly selected number of observations in the range $[1,$ min(segment length $-100$)$]$, (2) randomly assigning a wrong correlation structure pattern to a randomly increasing number of segments in the range $[1,$ max(n segments)$]$, and (3) combining (1) and (2). We ensure that these degraded quality clusterings cover the full range of the Jaccard index ($[0,1]$). Additionally, we generated four more versions: two with reduced cluster counts (50\% and 25\% of original) and two with reduced segment counts (50\% and 25\% of original), to analyse how these reductions impact validation measures.

For reproducibility of data generation, we use fixed random seeds. To generate each subject, the main seed is consistently varied, ensuring reproducible but independent generation. The main seed for generating the exploratory dataset was $666$, respectively $1905$ for the confirmatory dataset. The seed used to generate the partial and sparse variants was $1661$ for the exploratory dataset, respectively $99$ for the confirmatory dataset. Given that each subject's complete data variant is independent, we do not vary the seeds for sparsification between subjects. To create a simulated degraded clustering, the seed used for the exploratory dataset was $666$, respectively $2122$ for the confirmatory dataset.

In total, \dbname{} contains $48960$ ($30$ subjects $\times 68$ clusterings (including ground truth) $\times 12$ data variants $\times 2$ datasets) pregenerated clustering results of which $720$ ($30$ subjects $\times 12$ data variants $\times 2$ datasets) are perfect ground truth clusterings (times four if we include the cluster and segment count reduced versions). For the complete data variants, each subject contains approximately $1.26$ million observations, providing substantial data for robust algorithm evaluation. More details on the characteristics and visualisations of the dataset can be found in the Appendix~\ref{app:dataset-description}.

\section{Benchmark Validation}\label{sec:benchmark-validation}
We validated \dbname{} to confirm correlation structure preservation across variants and quantify how distribution shifts, sparsification, and downsampling affected these structures. Our analysis revealed four key findings: (1) downsampling data from 1 second to 1 minute moderately distorts correlation structures; (2) distribution shifts and sparsification have minimal impact on structure preservation; (3) negative correlations are more vulnerable to distortion than positive ones; and (4) Spearman's correlation consistently outperforms alternatives, requiring at least $30$ observations per segment for acceptable estimation accuracy. To quantify correlation structure preservation across data variants, we measured (1) mean absolute error (MAE) between the relaxed target correlation structure with coefficients $p_i$ and their empirical estimates $a_i$ ($\text{MAE} = \frac{1}{n} \sum_{i=1}^{n} |p_i - a_i|$, $n=3$), and (2) the percentage of segments with correlation estimates outside the tolerance bands $\toleranceBands$ (see Section~\ref{sec:benchmark-design}). Lower MAE and fewer out-of-tolerance segments indicate better structure modelling and preservation. All analyses used Spearman's correlation on $3000$ segments per data variant ($30$ subjects $\times 100$ segments) unless otherwise noted. For more details and results, see Appendices~\ref{app:dataset-description} to \ref{app:benchmark-reference-values}.

\paragraph{Impact of Downsampling} When downsampling non-normal data from 1 second to 1 minute correlation structures get distorted. The mean MAE for the complete variants increases from $0.02$ (SD $0.02$; non-normal) to $0.13$ (SD $0.08$; downsampled), while for the sparse variants it increases from $0.03$ (SD $0.03$, non-normal) to $0.11$ (SD $0.07$, downsampled). In practical terms, this means that correlation patterns in the downsampled variants might no longer model strong positive or negative correlations and instead become moderately positive or negative correlated. For example, for the complete variants pattern 24 $[-0.71, -0.7, 0]$ (a pattern that is consistently less accurately modelled than other patterns) might have an empirical correlation structure of $[-0.67, -0.66, 0.04]$ (median MAE$=0.04$) in the non-normal variant, but become $[-0.53, -0.52, 0.18]$ (median MAE$=0.18$) in the downsampled variant.

\begin{figure}[h]
    \centering
    \includegraphics[width=0.9\textwidth]{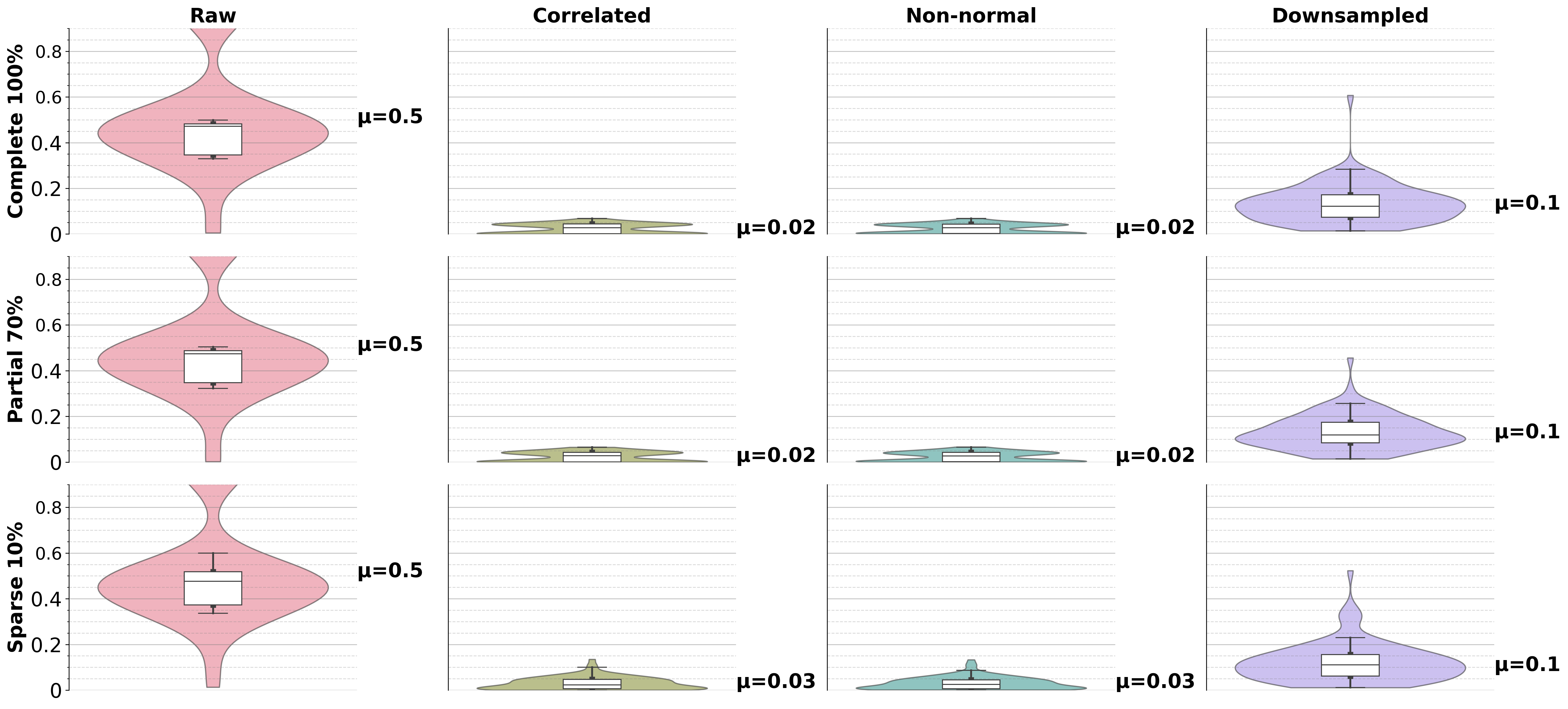}
    \caption{MAE distributions between target and empirical correlation structures across data variants. Lower values indicate better preservation of the original correlation structure.}
    \label{fig:mae-accross-data-variant}
\end{figure}

Further analysis of the effects of downsampling on different correlation structures for the complete data variant shows that pattern 13 $[1,1,1]$ with all positive correlations is the only pattern that remains unaffected by downsampling (MAE: $0.03$). Some simple structures with a single non-zero coefficient are less affected (e.g. pattern 9 $[1,0,0]$ and pattern 1 $[0,0,1]$ with MAE: $0.05$). Patterns containing negative correlations show the highest vulnerability to downsampling, with pattern 23 $[-1,1,-1]$ showing the highest degradation (MAE: $0.23$), followed by pattern 25 $[-1,-1,1]$ (MAE: $0.21$) and patterns 24 $[-0.71, -0.7, 0]$ and 18 $[-1,0,0]$ (MAE: $0.18$). Before downsampling, MAE values are similar across all patterns (range $0-0.04$). The upper end of this MAE range for these non-normal variants is for correlation structures that need relaxation to become a valid correlation structure, whereas the lower end is for correlation structures with a Hamming distance of $3$. Negative correlation coefficients do not inherently lead to higher MAE before downsampling.

\paragraph{Impact of Distribution Shifting} Distribution shifting from normal to non-normal distributions has minimal impact on the preservation of the correlation structures (MAE $0.03$ for all normal and non-normal data variants). Similarly, the number of segments outside of tolerance increases minimally (remains $1$ for complete, increases from $2$ to $3$ for partial, and from $13$ to $15$ for sparse data variants). Figure~\ref{fig:mae-accross-data-variant} shows that there are no visible differences in the correlation structures between the normal and non-normal data variants.

\paragraph{Impact of Sparsification} 
Sparsification to 10\% of the original observations preserves correlation structures remarkably well. Although the median MAE remains unaffected, the mean MAE increases from $0.2-0.3$ across both normal and non-normal variants. The effect on segments outside tolerance is minimal but noticeable, with median values increasing from $1$ to $3$ for 70\% sparsification and to $15$ for 10\% sparsification in non-normal variants.

\begin{figure}[htbp]
    \centering
    \includegraphics[width=0.9\textwidth]{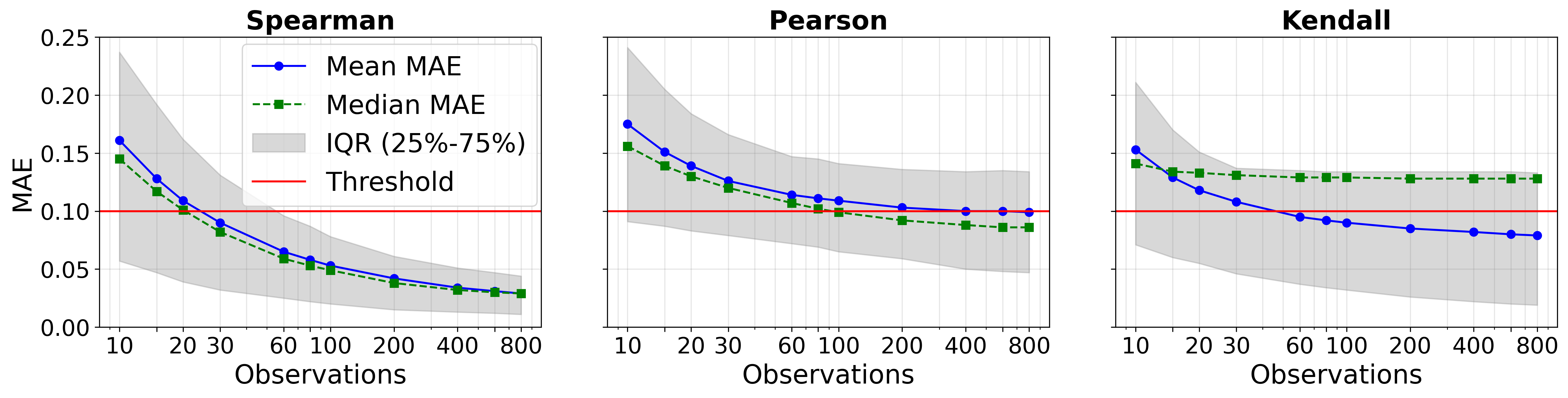}
    \caption{Effect of segment length on MAE between specified correlation structures and their estimation using different measures for the complete, non-normal data variant.}
    \label{fig:min-segment-length}
\end{figure}

\paragraph{Correlation Measures} Comparing MAE for different correlation measures reveals Spearman correlation consistently outperforms alternatives. For the normal data variants, Spearman ($0.03$) and Pearson ($0.04$) perform similarly, both outperforming Kendall ($0.14$). For the non-normal data variants, Spearman ($0.03$) outperforms Pearson ($0.08-0.09$) and Kendall ($0.13$). Spearman correlation coefficients become reliable for segments with at least $30$ observations (MAE $<0.1$), with $75\%$ of segments achieving MAE $<0.1$ at $60$ observations (Figure~\ref{fig:min-segment-length}). Adding more observations does not improve the results for the Pearson or Kendall correlation. The correlation structure distortion in downsampled data cannot be attributed to segment length limitations, as the mean segment length is $210$ observations (SD $185$, range $13-600$), more details see Appendix~\ref{app:correlation-measures}.

\paragraph{Exploratory and confirmatory datasets} We validated that the exploratory and confirmatory datasets maintain statistical equivalence while ensuring independence. Both datasets show nearly identical means and identical medians across key measures (MAE, segment length, correlation structures, and time gaps). Independence is confirmed by low Spearman correlation coefficients (all $<0.021$) with non-significant p-values (all $p > 0.25$) between the corresponding measures across datasets. Although both datasets contain the same number of subjects and use identical correlation structures, they vary independently in pattern ordering, segment lengths, and observation irregularities, making them suitable for two-phase statistical validation. For complete details, see Appendix Table~\ref{tab:dataset-independence-and-equivalence}.

\section{Benchmark Usage}\label{sec:benchmark-usage}
\subsection{Dataset Access} 
\dbname{} is available on Hugging Face (\href{https://huggingface.co/datasets/idegen/csts}{idegen/csts}). The dataset can be loaded using the Hugging Face Datasets library, as shown in code snippet~\ref{lst:hf_dataset_loading}. The documentation on Hugging Face provides detailed instructions for accessing the exploratory and confirmatory dataset splits for all data variants, with a linked Google Colabratory notebook that demonstrates how to work with each subset of the dataset.
\begin{lstlisting}[caption={Loading data and ground truth labels for the complete correlated data variant.},label={lst:hf_dataset_loading}]
from datasets import load_dataset
split = "exploratory"
config_data = "correlated_complete_data"
data = load_dataset("idegen/csts", name=config_data, split=split)

config_labels = "correlated_complete_labels"
labels = load_dataset("idegen/csts", name=, split=split)
\end{lstlisting}

\subsection{Key Research Applications}
\dbname{}'s enables rigorous evaluation in the following research areas: correlation-based clustering algorithms, preprocessing techniques, and validation methods.

\paragraph{Evaluating algorithms and preprocessing effects}
Researchers can evaluate time series clustering algorithms across different data variants to isolate specific sensitivities, such as distribution sensitivity by comparing performance between normal and non-normal data variants, or robustness to sparsity by evaluating across completeness levels. \dbname{} facilitates the analysis of the effects of preprocessing on correlation structures, as we have demonstrated with downsampling. Researchers can apply other preprocessing techniques and compare the results with our benchmark results to make an informed decision about the effectiveness of these techniques for correlation-based clustering. Our comprehensive evaluation framework available through GitHub provides standardised measures, cluster result mapping utilities, and visualisation tools to systematically assess algorithm performance and compare with our established threshold values. With its controlled properties and comprehensive ground truth labels, \dbname{} bridges the gap between theoretical models and real-world data where ground truth labels are not available.

\paragraph{Evaluating clustering validation methods}
\dbname{} supports the evaluation of clustering validation methods for correlation-based structures. The dataset includes controlled degraded segmentation and clustering results that span the entire Jaccard index range, allowing researchers to evaluate novel validation methods with clear performance thresholds. Our analysis has established validated thresholds for strong correlation structures (silhouette width coefficient $>0.8$, Davies-Bouldin index $<0.2$), providing clear reference points for interpreting clustering quality; see Appendix~\ref{app:benchmark-reference-values}.

\subsection{Evaluation Protocol}
We recommend the following standardised protocol for consistent and comparable assessments:

\paragraph{Cluster-to-Ground-Truth Mapping}
Map algorithm results to \dbname{}'s ground truth correlation structures by calculating the Spearman correlation of all observations in each algorithm-discovered cluster and matching this empirical cluster structure to the closest ground truth structure(s) using the L1 norm distance.

\paragraph{Performance Evaluation} 
Assess structural quality by evaluating internal indices (Silhouette Width Coefficient (SWC) \cite{Rousseeuw1987} and Davies-Bouldin Index (DBI) \cite{Davies1979}) using the L5 norm distance, and the Jaccard index \cite{Vendramin2010} as external validation. Evaluate pattern accuracy by calculating MAE between each segment's empirical correlation and its mapped ground truth correlation structure, pattern discovery rate (percentage of ground truth correlation structures matched at least once within tolerance $\pm0.1$, see Section~\ref{sec:benchmark-design}), and pattern specificity (percentage of algorithm clusters matching exactly one ground truth cluster within tolerance $\pm0.1$). Quantify segmentation quality through segment count and length ratios compared to ground truth.

\paragraph{Results interpretation}
Use the Tables in Appendix~\ref{app:benchmark-reference-values} that show both the ground truth baseline and the systematically degraded results to contextualise algorithm performance. Consider lower MAE values as an indication of a closer alignment between the discovered and ground truth correlation structures, and SWC $>0.8$ and DBI $<0.2$ as indicators of good structural quality of the clustering. Interpret higher pattern discovery percentages as evidence of more patterns discovered; higher pattern specificity percentages as better algorithm-to-ground truth matching; segmentation ratio values $>1$ as over-segmentation; and segment length ratio values $>1$ as excessive segment length.

\paragraph{Statistical Validation}
Confirm that the differences in results between data variants or techniques are significant by applying Wilcoxon signed rank tests on the results of the paired subjects. The structure of the dataset supports exploratory analysis on the exploratory split and independent confirmation of significant findings using the confirmatory split.

\subsection{Extension}
Beyond the pre-generated datasets, our generation framework supports customisation through the GitHub repository. To match their domain-specific data properties, researchers can modify correlation structures, adjust distribution properties, change segment lengths, downsampling rate, sparsity, change the number of subjects and time series variates generated.

\subsection{Reproducibility}
Our data generation framework, evaluation tools, documentation, and Conda-configured environment setup are available at \url{https://github.com/isabelladegen/corrclust-validation} for reproducibility. Data generation, validation, and experiments were run on standard hardware (MacBook Pro with an Apple 8-core M1 chip and 16GB RAM).

\section{Case Study: Time Series Clustering Algorithm Evaluation}\label{sec:case-study}
We demonstrate \dbname{}'s utility by evaluating Toeplitz Inverse Covariance-based Clustering (TICC) \cite{Hallac2017}, an algorithm designed for segmentation and clustering of time series data. TICC models clusters using Gaussian inverse covariance matrices to discover correlation structures in temporal data. Although TICC was previously tested on synthetic and proprietary real-world sensor data, its sensitivities to non-normal distributions and sampling irregularities remained unevaluated. We applied untuned TICC (clusters$=23$, window$=5$, switch penalty$=400$, lambda$=0.11$, max iterations$=10$) to our benchmark. We trained on one exploratory subject and applied the model to the remaining $29$ subjects. Full experimental details and results are provided in the Appendix~\ref{appendix:use-case-experiments}.

Table~\ref{tab:case-study-ticc} reveals performance disparities between normal and non-normal data variants. For normal data, TICC achieved reasonable performance with high pattern discovery rates ($\sim80\%$) and specificity ($>87\%$) across all completeness levels. TICC did not achieve our recommended performance measures of SWC$>0.8$ and DBI$<0.2$. Comparison with controlled degraded validation measures indicates that TICC's results are comparable with $\sim5$ segments assigned to an incorrect cluster or $200-400$ observations missing from each segment. Our recommendation would be to try to improve TICC's performance by tuning the hyperparameters. In contrast, with non-normal variants, TICC's performance deteriorates dramatically across all performance measures. Internal validation measures (negative SWC, extremely high DBI) confirm that TICC failed to identify the correlation structures, revealing its distribution sensitivity not examined in the original publication, suggesting that TICC requires a normalising preprocessing step.

\begin{table}[!ht]
\caption{Performance measures (means) of TICC compared to \dbname{} ground truth (GT) for normal and non-normal data variants. More results are provided in Appendix~\ref{appendix:use-case-experiments}.}
\label{tab:case-study-ticc}
\setlength{\tabcolsep}{3.8pt}
\renewcommand{\arraystretch}{1.2} 
\small
\centering
\begin{tabular}{lcc|cc|cc|cc|cc|cc}
\toprule
& \multicolumn{6}{c|}{\textbf{Normal}} & \multicolumn{6}{c}{\textbf{Non-normal}} \\
\cmidrule{2-13}
\textbf{Completeness} & \multicolumn{2}{c|}{\textbf{100\%}} & \multicolumn{2}{c|}{\textbf{70\%}} & \multicolumn{2}{c|}{\textbf{10\%}} & \multicolumn{2}{c|}{\textbf{100\%}} & \multicolumn{2}{c|}{\textbf{70\%}} & \multicolumn{2}{c}{\textbf{10\%}} \\
\midrule
\textbf{Measures} & \textbf{TICC} & \textbf{GT} & \textbf{TICC} & \textbf{GT} & \textbf{TICC} & \textbf{GT} & \textbf{TICC} & \textbf{GT} & \textbf{TICC} & \textbf{GT} & \textbf{TICC} & \textbf{GT} \\
\midrule
SWC & 0.73 & 0.97 & 0.72 & 0.97 & 0.61 & 0.92 & -0.15 & 0.97 & -0.15 & 0.97 & -0.08 & 0.92 \\
DBI & 0.74 & 0.04 & 0.54 & 0.05 & 1.04 & 0.14 & 2.85 & 0.04 & 3.59 & 0.04 & $\ast$ & 0.14 \\
Jaccard & 0.82 & 1 & 0.79 & 1 & 0.80 & 1 & 0.38 & 1 & 0.38 & 1 & 0.28 & 1 \\
MAE & 0.07 & 0.02 & 0.05 & 0.02 & 0.07 & 0.02 & 0.15 & 0.02 & 0.15 & 0.02 & 0.14 & 0.03 \\
Pattern Discovery& 81.2 & 100 & 78.4 & 100 & 78.4 & 100 & 49.6 & 100 & 50.3 & 100 & 37.5 & 100 \\
Pattern Specificity& 91 & 100 & 98.4 & 100 & 87.4 & 100 & 54.2 & 100 & 56.8 & 100 & 63.2 & 100 \\
\bottomrule
\multicolumn{13}{l}{\footnotesize $\ast$ DBI for 10\% Non-normal: $>$19Mio (due to division $\sim0$ for similar cluster centroids), min 1.17}
\end{tabular}
\end{table}

TICC performed better for complete data, where it sometimes missed only $2-3$ patterns. As data sparsity increases, the consistency of pattern detection decreases. Our analysis showed that TICC consistently struggled with specific correlation structures despite their accurate representation in each data variant. It performed better with simple correlation structures (particularly pattern 0 $[0,0,0]$, pattern 13 $[1,1,1]$, and patterns with single strong correlations).

Our case study demonstrates \dbname{}'s capabilities for: (1) quantifying algorithm robustness across data conditions, (2) identifying TICC's undocumented distribution sensitivity, (3) separating algorithm limitations from data quality issues, and (4) enabling objective benchmarking and hyperparameter tuning that would be impossible in real-world clustering scenarios due to lacking ground truth.

\section{Limitations}\label{sec:limitations}
\dbname{} has several important limitations. The dataset models correlation structures that change between segments, resulting in data lacking temporal dependencies (autocorrelation, trends, seasonality) found in typical time series. Instead, \dbname{} focusses on stationary segments with regime-switching correlation structures.
Our approach is limited to modelling correlation structures between three time series variates, yielding $23$ distinct, interpretable correlation structures. This constraint is deliberate, as extending to four variates would exponentially increase the correlation structure space to $729$ possibilities, which is beyond human ability to assign meaningful interpretation to each pattern. Furthermore, we explore specific distributions (normal, extreme value, negative binomial) within set parameter ranges, sparsity levels ($100\%$, $70\%$, $10\%$), and segment lengths ($15$ min to $12$ hours). However, \dbname{}'s synthetic data generation framework is highly configurable, allowing researchers to generate their own data that better suit their needs with regard to number of variates, distribution configurations, sparsity levels, segment lengths, subject counts, and even alternative correlation patterns.

\section{Conclusion}\label{sec:conclusion}
We introduced \dbname{}, a comprehensive correlation structure-specific benchmark for time series clustering with validated ground truth across controlled data variations. Our benchmark enables rigorous assessment of both clustering algorithms and validation methods while clearly distinguishing between algorithmic limitations and structure degradation. Key empirical findings reveal that correlation structures are minimally affected by the distribution shifts and sparsification considered, while downsampling can weaken strong correlation structures into moderate ones. Negative correlations are more vulnerable to distortion than positive ones, Spearman correlation consistently outperforms alternatives, and reliable estimation requires at least $30$ observations per segment for a correlation coefficient estimation with an MAE $<0.1$.

Based on these findings, we advise that researchers maintain high-frequency sampling for accurate correlation estimation and develop algorithms that work directly with irregular data. We recommend caution when downsampling, as correlation structures (especially those with negative correlations) can be distorted even with adequate observations. Researchers should ensure that the target real-world application contains enough observations per segment and use Spearman correlation estimation. With its validated correlation structures, standardised evaluation protocols, and extensible data generation framework, \dbname{} represents a shift towards structure-first clustering evaluation, where benchmarks are tied to specific structural properties rather than arbitrary classification boundaries. Although we focus on correlation structures, our approach provides a template for establishing rigorous benchmarks for other fundamental data structures.

\section*{Acknowledgements} 
We would like to thank UK Research and Innovation (UKRI) for funding author ID's PhD research through the UKRI Doctoral Training in Interactive Artificial Intelligence (AI) under grant EP/S022937/1. The authors extend their gratitude to the faculty, staff and colleagues of the Interactive AI Centre for Doctoral Training at Bristol University for their valuable support and guidance throughout this research.
We acknowledge the use of Claude 3.7 Sonnet by Anthropic as a research dialogue tool throughout the development of this work, assisting with dataset documentation, iterative refinement of ideas, and evaluating the clarity of our methods and contributions.

\bibliographystyle{unsrturl}  
\bibliography{main} 
\clearpage


\appendix
\section{Dataset Characteristics}\label{app:dataset-description}

\subsection{Key statistics}
This section provides descriptive statistics for both the exploratory and the confirmatory datasets across all data variants. Tables~\ref{tab:dataset-key-statistics} and \ref{tab:dataset-key-statistics-confirmatory} summarise the mean absolute error (MAE) between the relaxed target correlation structure and the empirical correlation structure, the number of segments outside the tolerance bands $\toleranceBands$ (see Section~\ref{sec:benchmark-design}), the lengths of the segments, and the observations counts, highlighting the differences between the data variants. MAE and segment length are averaged across segments ($100$ segments for each of the $30$ subjects), while segments outside of tolerance and observation count are averaged across the $30$ subjects for each variant.

\begin{table*}[!h]
\caption{Descriptive statistics of \textbf{exploratory dataset} for mean average error (MAE) between empirical correlations and target relaxed canonical patterns, number of segments outside tolerance bands, segment lengths and observation counts for each data variant showing differences between data completeness (rows) and generation stages (columns). }
\label{tab:dataset-key-statistics}
\small
\setlength{\tabcolsep}{4pt}
\centering
\begin{tabular*}{\textwidth}{@{\extracolsep{\fill}}l l l cccc}
\toprule
\multirow{2}{*}{\textbf{Completeness}} &
  \multicolumn{2}{c}{\multirow{2}{*}{\textbf{Descriptive Statistics}}} &
  \multicolumn{4}{c}{\textbf{Generation Stages}} \\
 & & & \textbf{Raw} & \textbf{Correlated} & \textbf{Non-normal} & \textbf{Downsampled} \\
\midrule
\multirow{9}{*}{\textbf{Complete} (100\%)} & \multirow{2}{*}{Correlation MAE} & $\mu$ (SD) & 0.51 (0.25) & 0.02 (0.02) & 0.02 (0.02) & 0.13 (0.08) \\
 & & range & 0.001 - 1.04 & 0 - 0.09 & 0 - 0.09 & 0.004 - 0.61 \\
\cmidrule(l){2-7}
& Segments outside & $\mu$ (SD) & 95.6 (0.5) & 1.9 (1.16) & 4.23 (8.49) & 67.6 (7.24) \\
 &tolerance & range & 95 - 96 & 0 - 4 & 0 - 48 & 55 - 80 \\
\cmidrule(l){2-7}
 & \multirow{2}{*}{Segment lengths} & $\mu$ (SD) & \multicolumn{3}{c}{12640.1 (11118.7)} & 210.7 (185.3) \\
 & & range & \multicolumn{3}{c}{900 - 36000} & 15 - 600 \\
\cmidrule(l){2-7}
 & \multirow{2}{*}{Observation count} & $\mu$ (SD) & \multicolumn{3}{c}{1264010 (11325.6)} & 21066.8 (188.8) \\
 & & range & \multicolumn{3}{c}{1243800 - 1284300} & 20730 - 21405 \\
\midrule
\multirow{9}{*}{\textbf{Partial} (70\%)} & \multirow{2}{*}{Correlation MAE} & $\mu$ (SD) & 0.51 (0.25) & 0.03 (0.02) & 0.03 (0.02) & 0.13 (0.08) \\
& & range & 0.001 - 1.04 & 0 - 0.09 & 0 - 0.09 & 0.0 - 0.56 \\
\cmidrule(l){2-7}
& Segments outside & $\mu$ (SD) & 95.6 (0.5) & 3.07 (1.70) & 6.03 (8.29) & 67.2 (7.06) \\
 &tolerance & range & 95 - 96 & 0 - 6 & 1 - 48 & 55 - 79 \\
\cmidrule(l){2-7}
 & \multirow{2}{*}{Segment lengths} & $\mu$ (SD) & \multicolumn{3}{c}{8848.1 (7783.2)} & 210.7 (185.3) \\
 & & range & \multicolumn{3}{c}{592 - 25425} & 15 - 600 \\
\cmidrule(l){2-7}
 & \multirow{2}{*}{Observation count} & $\mu$ (SD) & \multicolumn{3}{c}{884807 (7927.9)} & 21066.8 (188.8) \\
 & & range & \multicolumn{3}{c}{870660 - 899010} & 20730 - 21405 \\
\midrule
\multirow{9}{*}{\textbf{Sparse} (10\%)} & \multirow{2}{*}{Correlation MAE}  & $\mu$ (SD) & 0.52 (0.24) & 0.03 (0.03) & 0.03 (0.03) & 0.11 (0.07) \\
& & range & 0.002 - 1.14 & 0 - 0.19 & 0 - 0.19 & 0.0 - 0.52\\
\cmidrule(l){2-7}
& Segments outside & $\mu$ (SD) & 95.8 (0.61) & 14.6 (2.58) & 18.07 (6.15) & 61.7 (5.29) \\
 &tolerance & range & 95 - 97 & 10 - 20 & 12 - 46 & 54 - 74 \\
\cmidrule(l){2-7}
 & \multirow{2}{*}{Segment lengths} & $\mu$ (SD) & \multicolumn{3}{c}{1264 (1112.8)} & 210.3 (185) \\
 & & range & \multicolumn{3}{c}{58 - 3771} & 13 - 600 \\
\cmidrule(l){2-7}
 & \multirow{2}{*}{Observation count} & $\mu$ (SD) & \multicolumn{3}{c}{126401 (1132.6)} & 21030.2 (187.7) \\
 & & range & \multicolumn{3}{c}{124380 - 128430} & 20702 - 21368 \\
\bottomrule
\end{tabular*}
\end{table*}

\begin{table*}[!h]
\caption{Descriptive statistics of \textbf{confirmatory dataset} for mean average error (MAE) between empirical correlations and target relaxed canonical patterns, number of segments outside tolerance bands, segment lengths and observation counts for each data variant showing differences between data completeness (rows) and generation stages (columns). }
\label{tab:dataset-key-statistics-confirmatory}
\small
\setlength{\tabcolsep}{4pt}
\centering
\begin{tabular*}{\textwidth}{@{\extracolsep{\fill}}l l l cccc}
\toprule
\multirow{2}{*}{\textbf{Completeness}} &
  \multicolumn{2}{c}{\multirow{2}{*}{\textbf{Descriptive Statistics}}} &
  \multicolumn{4}{c}{\textbf{Generation Stages}} \\
 & & & \textbf{Raw} & \textbf{Correlated} & \textbf{Non-normal} & \textbf{Downsampled} \\
\midrule
\multirow{9}{*}{\textbf{Complete} (100\%)} & \multirow{2}{*}{Correlation MAE} & $\mu$ (SD) & 0.51 (0.25) & 0.03 (0.02) & 0.03 (0.02) & 0.13 (0.08) \\
 & & range & 0 - 1.03 & 0 - 0.1 & 0 - 0.1 & 0 - 0.53 \\
\cmidrule(l){2-7}
& Segments outside & $\mu$ (SD) & 95.7 (0.48) & 1.6 (0.97) & 4.3 (9,07) & 68.6 (7.13)\\
 &tolerance & range & 95 - 96 & 0 - 4 & 0 - 51 & 54 - 83\\
\cmidrule(l){2-7}
 & \multirow{2}{*}{Segment lengths} & $\mu$ (SD) & \multicolumn{3}{c}{12652.2 (11137.9)} & 210.9 (185.6) \\
 & & range & \multicolumn{3}{c}{900 - 36000} & 15 - 600 \\
\cmidrule(l){2-7}
 & \multirow{2}{*}{Observation count} & $\mu$ (SD) & \multicolumn{3}{c}{1265220 (11348.3)} & 21087 (189.1)\\
 & & range & \multicolumn{3}{c}{1239600 - 1291500} & 20660 - 21525 \\
\midrule
\multirow{9}{*}{\textbf{Partial} (70\%)} & \multirow{2}{*}{Correlation MAE} & $\mu$ (SD) & 0.51 (0.25) & 0.03 (0.02) & 0.03 (0.02) & 0.13 (0.08)\\
& & range & 0 - 1.05 & 0 - 0.1 & 0 - 0.1 & 0 - 0.58\\
\cmidrule(l){2-7}
& Segments outside & $\mu$ (SD) & 95.7 (0.48) & 2.7 (1.58) & 5.7 (9.04) & 68.7 (6.65)\\
 &tolerance & range & 95 - 96 & 0 - 6  & 0 - 51 & 55 - 80\\
\cmidrule(l){2-7}
 & \multirow{2}{*}{Segment lengths} & $\mu$ (SD) & \multicolumn{3}{c}{8856.5 (7796.1)} & 210.9 (185.6) \\
 & & range & \multicolumn{3}{c}{584 - 25421} & 15 - 600 \\
\cmidrule(l){2-7}
 & \multirow{2}{*}{Observation count} & $\mu$ (SD) & \multicolumn{3}{c}{885654 (7943.8)} & 21087 (189.1) \\
 & & range & \multicolumn{3}{c}{867720 - 904050} & 20660 - 21525 \\
\midrule
\multirow{9}{*}{\textbf{Sparse} (10\%)} & \multirow{2}{*}{Correlation MAE}  & $\mu$ (SD) & 0.52 (0.24) & 0.03 (0.03) & 0.03 (0.03) & 0.11 (0.07)\\
& & range & 0.003 - 1.17 & 0 - 0.24 & 0 - 0.24 & 0 - 0.47\\
\cmidrule(l){2-7}
& Segments outside & $\mu$ (SD) & 95.8 (0.5) & 14.9 (3.75) & 17.6 (7.13) & 62.2 (5.05)\\
 &tolerance & range & 95 - 97 & 9 - 23 & 9 - 46 & 54 - 75\\
\cmidrule(l){2-7}
 & \multirow{2}{*}{Segment lengths} & $\mu$ (SD) & \multicolumn{3}{c}{1265.2 (1114.9)} & 210.5 (185.3) \\
 & & range & \multicolumn{3}{c}{67 - 3760} & 14 - 600 \\
\cmidrule(l){2-7}
 & \multirow{2}{*}{Observation count} & $\mu$ (SD) & \multicolumn{3}{c}{126522 (1134.8)} & 21049.2 (190.4) \\
 & & range & \multicolumn{3}{c}{123960 - 129150} & 20623 - 21489 \\
\bottomrule
\end{tabular*}
\end{table*}

\subsection{Distribution Description}
\begin{figure}[h]
\centerline{\includegraphics[width=\textwidth]{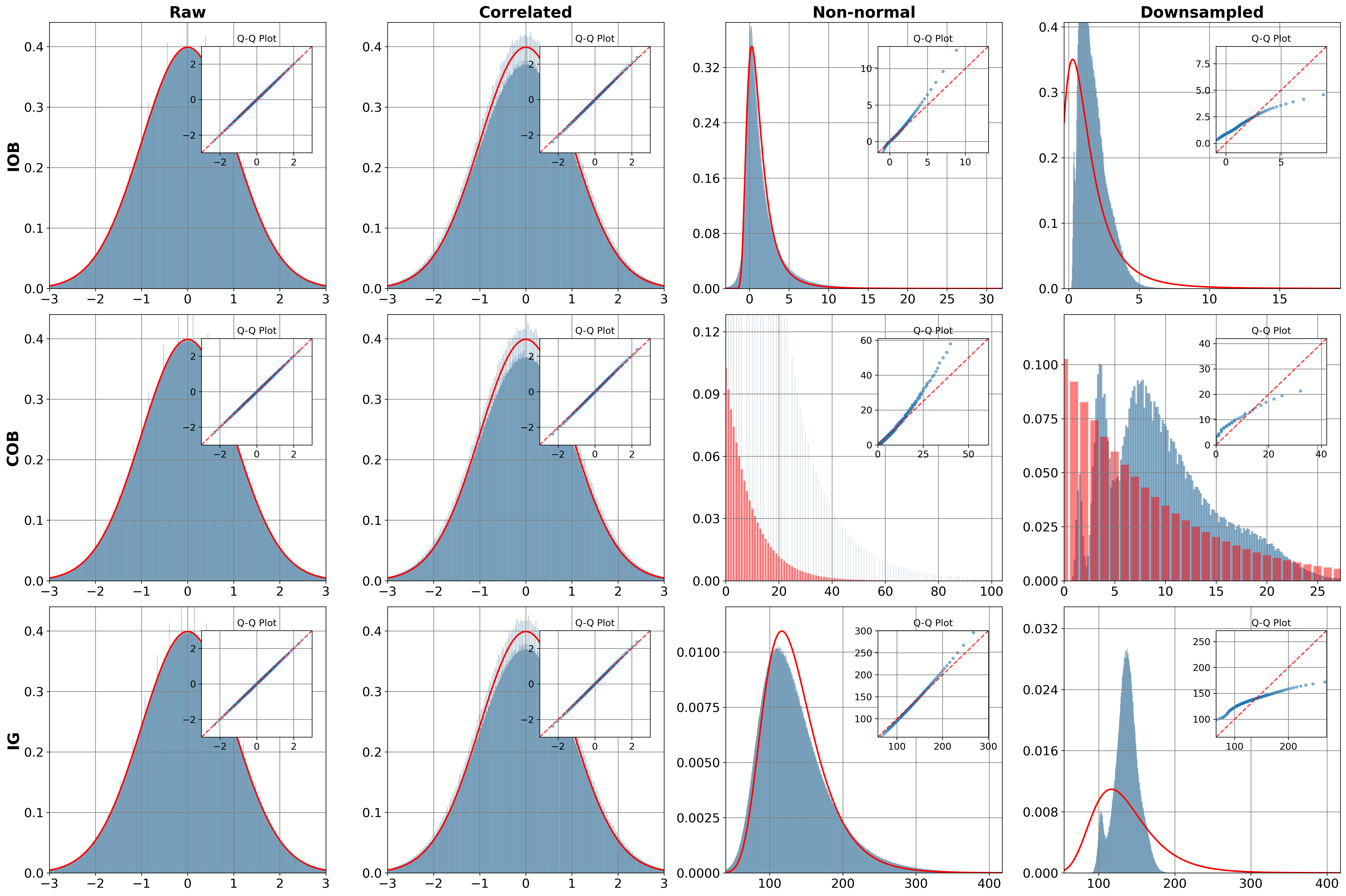}}
    \centering
    \caption{Empirical distributions (blue) with theoretical PDF/PMF (red) for the \textbf{complete variants} (columns) and the three time series variates (rows). Q-Q plots in insets show quantile comparison between empirical and theoretical distributions. Raw and correlated variants show the standard normal distribution as theoretical distribution, non-normal and downsampled variants show the median parameters of the non-normal distributions.}
    \label{fig:distribution-complete}
\end{figure}

\begin{figure}[h]
\centerline{\includegraphics[width=\textwidth]{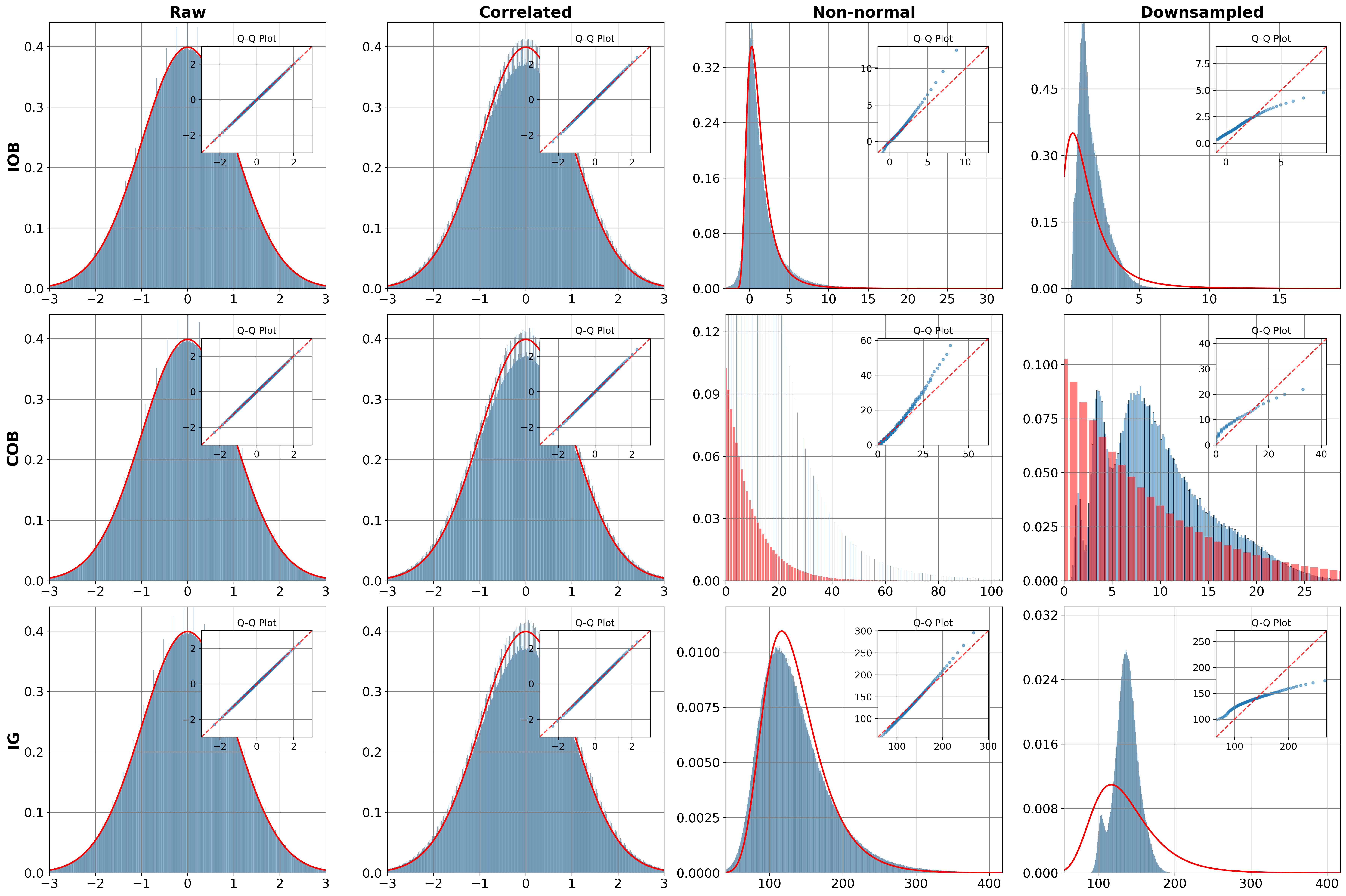}}
    \centering
    \caption{Empirical distributions (blue) with theoretical PDF/PMF (red) for the \textbf{partial variants} (70\% observations) (columns) and the three time series variates (rows). Q-Q plots in insets show quantile comparison between empirical and theoretical distributions. Raw and correlated variants show the standard normal distribution as theoretical distribution, non-normal and downsampled variants show the median parameters of the non-normal distributions.}
    \label{fig:distribution-partial}
\end{figure}

\begin{figure}[h]
\centerline{\includegraphics[width=\textwidth]{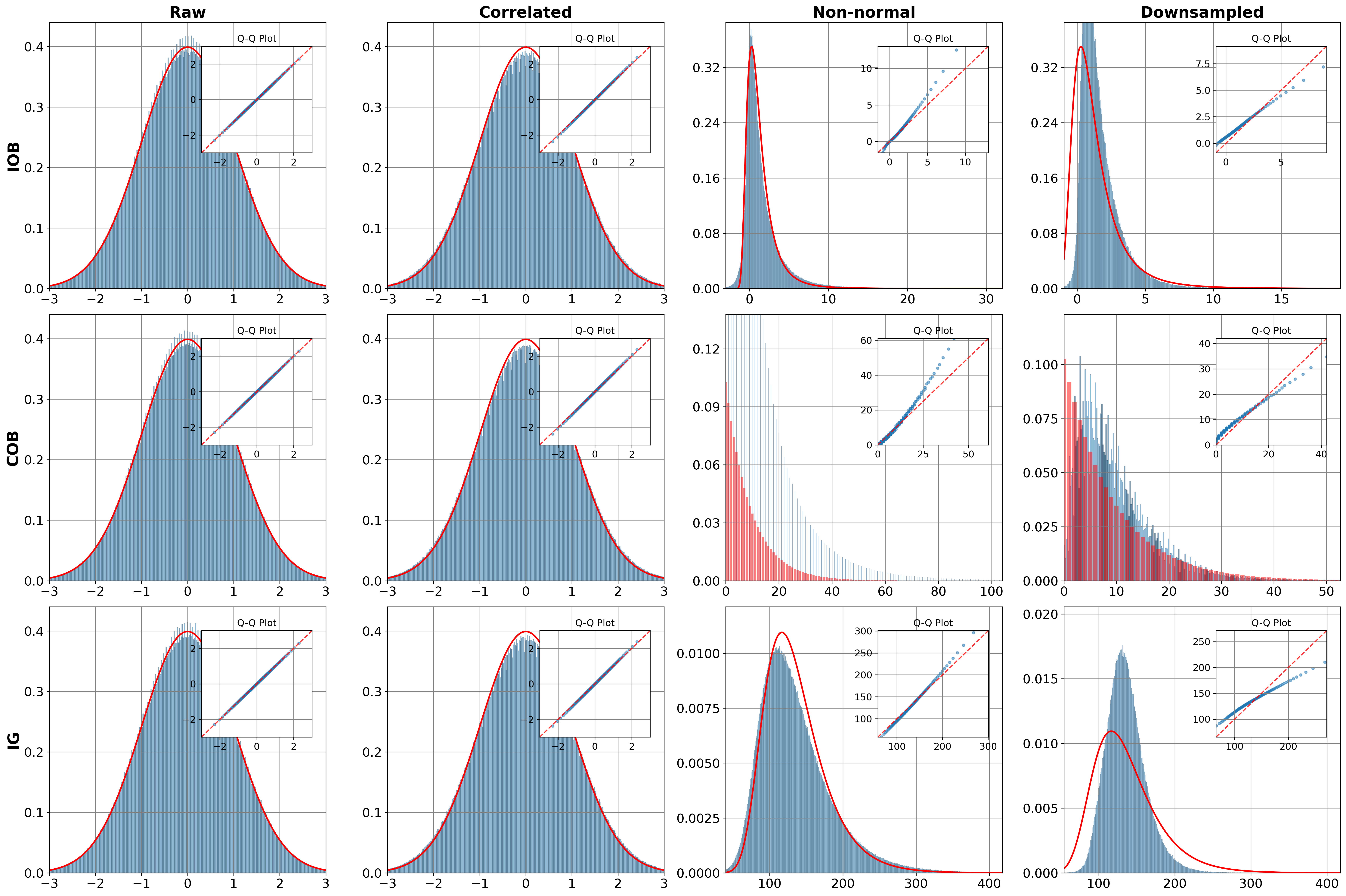}}
    \centering
    \caption{Empirical distributions (blue) with theoretical PDF/PMF (red) for the \textbf{sparse variants} (10\% observations) (columns) and the three time series variates (rows). Q-Q plots in insets show quantile comparison between empirical and theoretical distributions. Raw and correlated variants show the standard normal distribution as theoretical distribution, non-normal and downsampled variants show the median parameters of the non-normal distributions.}
    \label{fig:distribution-sparse}
\end{figure}

Figures~\ref{fig:distribution-complete}-\ref{fig:distribution-sparse} illustrate the distribution characteristics across the four data generation stages (raw, correlated, non-normal, and downsampled) for each of the three time series variates (IOB, COB, and IG). Each subplot contains an empirical histogram (blue) with the theoretical probability density function (PDF) or probability mass function (PMF) overlaid (red line), along with a Q-Q plot inset that compares empirical quantiles against theoretical quantiles. For the raw and correlated data variants, a standard normal distribution ($\mu=0$, $\sigma=1$) provides the theoretical reference, reflecting the original data generation process. For the non-normal and downsampled data variants, the theoretical distributions are specific to the each variate using median parameters: an extreme value distribution for IOB (top row), a negative binomial distribution for COB (middle row), and an extreme value distribution for IG (bottom row).

The visualisations show that (1) the raw and correlated variants follow a standard normal distribution, confirming that correlating the variates preserves the distributions of the observations; (2) the distribution shifting for the non-normal variants is successful; and (3) the downsampled variants (60 observations mean aggregated into one) preserves the non-normal distributions with some distortion confirming that the time series variates (IOB, COB, IG) in the non-normal data variants are not independent.

\subsection{Sparsification Description}

\begin{figure}[h]
\centerline{\includegraphics[width=0.9\textwidth]{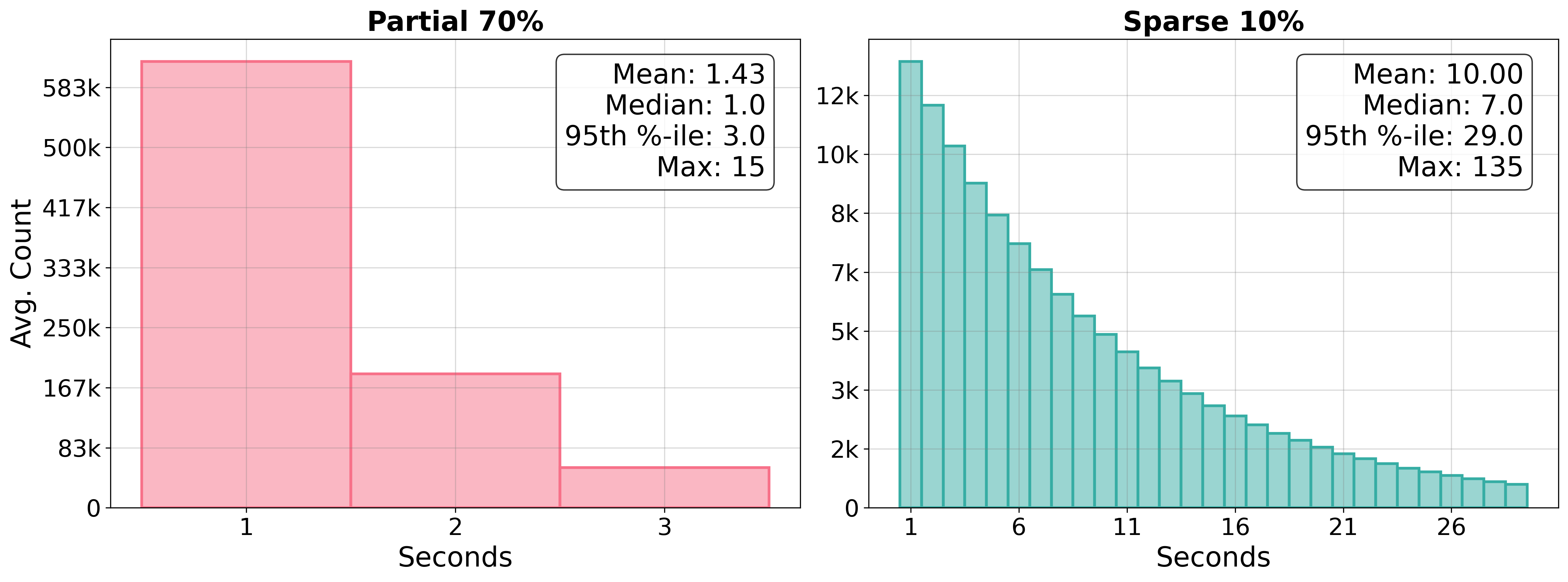}}
    \centering
    \caption{Histogram of time intervals between observations (in second) for the \textbf{partial and sparse non-normal data variants}}
    \label{fig:time-gaps-partial-sparse}
\end{figure}

Figure~\ref{fig:time-gaps-partial-sparse} illustrates the time intervals between consecutive observations for the irregular sampled non-normal data variants. The partial variant shows predominantly 1-second intervals (mean: $1.43$s, median: $1.0$s) with occasional gaps up to $15$ seconds. The sparse variant shows larger gaps between observations (mean: $10.0$s, median: $7.0$s) that extend up to $135$ seconds. These distinct gap patterns demonstrate the varying degrees of temporal irregularity in the dataset, providing realistic conditions for evaluating the robustness of time series clustering algorithms to irregular sampling.

\subsection{Equivalence and Independence}
Table~\ref{tab:dataset-independence-and-equivalence} shows that the exploratory and confirmatory datasets maintain statistical equivalence while ensuring independence. To demonstrate independence, we calculate Spearman's r between the exploratory and confirmatory datasets for the relevant measures for each of the $30$ subjects and their $100$ segments. To demonstrate equivalence, we compare the mean, median, and IQR values for these values between the datasets. Relaxed MAE is calculated between the empirical correlation of each segment and the relaxed target pattern used to correlate the data for the segment. Pattern id refers to the id of a each of the canonical correlation patterns. The lack of correlation between exploratory and confirmatory relaxed MAEs, respectively, pattern IDs shows that the pattern for each segment was chosen at random. The lack of correlation between the lengths of the exploratory and confirmatory segments shows that the length of each segment was chosen at random. The time gaps are the time in seconds between the observations. The lack of correlations between the time gaps in the exploratory and confirmatory datasets shows that irregularities have been introduced at random for the partial (70\%) and sparse (10\%) data variants. These validations were run on the correlated data variant from which the non-normal and downsampled versions are generated. 
\begin{table}[!ht]
    \caption{Results showing that the exploratory (expl.) and the confirmatory (conf.) datasets are independent with low Spearman r correlations for various measures while being statistically equivalent with small differences in mean, and no differences in median and IQR.}
    \label{tab:dataset-independence-and-equivalence}
    \centering
  \begin{tabular}{l@{\hspace{6pt}}r@{\hspace{6pt}}r@{\hspace{6pt}}r@{\hspace{6pt}}r@{\hspace{6pt}}r@{\hspace{6pt}}r@{\hspace{6pt}}r@{\hspace{6pt}}r@{\hspace{6pt}}r@{\hspace{6pt}}r}
    \toprule
     \textbf{Measures, data} & \multicolumn{2}{c}{\textbf{Correlation}} & \multicolumn{2}{c}{\textbf{Mean}} & \multicolumn{2}{c}{\textbf{Median}} & \multicolumn{2}{c}{\textbf{25\%}} & \multicolumn{2}{c}{\textbf{75\%}} \\
    \textbf{completeness} & \textbf{r} & \textbf{p} & \textbf{Expl.} & \textbf{Conf.} & \textbf{Expl.} & \textbf{Conf.} & \textbf{Expl.} & \textbf{Conf.} & \textbf{Expl.} & \textbf{Conf.} \\
    \midrule
    Relaxed MAE, 100\% & 0.01 & 0.49 & 0.02 & 0.03 & 0.03 & 0.03 & 0.00 & 0.00 & 0.04 & 0.04 \\
    Segment length, 100\% & -0.02 & 0.25 & 12640 & 12652 & 10800 & 10800 & 1800 & 1800 & 18900 & 18900 \\
    Pattern ID, 100\% & 0.02 & 0.35 & 4.3 & 4.3 & 4 & 4 & 4 & 4 & 5 & 5 \\
    Time gaps, 70\% & 0.0 & 0.75 & 1.4 & 1.4 & 1 & 1 & 1 & 1 & 2 & 2 \\
    Time gaps, 10\% & 0.0 & 0.43 & 10.0 & 10.0 & 7 & 7 & 3 & 3 & 14 & 14 \\
    \bottomrule
  \end{tabular}
\end{table}

\clearpage

\section{Correlation structures}\label{app:correlation-structures}

\subsection{Canonical Pattern Catalogue}
Table~\ref{tab:cannonical-patterns} catalogues all $27$ theoretically possible correlation structures for three time series variates when modelling strong, negligible, and strong negative correlations. The catalogue identifies which can be modelled within our tolerance bands $\toleranceBands$. Of the $27$ possible patterns, $23$ can be modelled as valid positive semi-definite correlation matrices through appropriate relaxation. Patterns marked 'No' in the 'Ideal' column required coefficient relaxation within the tolerance bands $\toleranceBands={[-1,-0.7],[-0.2,0.2],[0.7,1]}$ to create valid correlation matrices, while patterns marked 'No' in the 'Modelled' column could not be represented as valid correlation matrices even with relaxation. The tolerance bands were defined to keep the relaxed coefficients clearly within the strong or negligible correlation coefficients while creating an equivalent threshold for all coefficients.
\begin{table}[!ht]  
\caption{Overview of all correlation structures and their relaxed versions. Indicating which patterns can be modelled exactly as they are and which need adjustment or cannot be modelled within the tolerance band.}
\label{tab:cannonical-patterns}
\begin{tabular*}{\columnwidth}{@{\extracolsep{\fill}}l cccc}
\toprule
\textbf{Id} & 
\textbf{Canonical Pattern} & 
\textbf{Relaxed Pattern} & 
\textbf{Ideal} & 
\textbf{Modelled} \\
\midrule
0 & (0, 0, 0) & (0, 0, 0) & Yes & Yes \\
1 & (0, 0, 1) & (0, 0, 1) & Yes & Yes \\
2 & (0, 0, -1) & (0, 0, -1) & Yes & Yes \\
3 & (0, 1, 0) & (0, 1, 0) & Yes & Yes \\
4 & (0, 1, 1) & (0.0, 0.71, 0.7) & No & Yes \\
5 & (0, 1, -1) & (0, 0.71, -0.7) & No & Yes \\
6 & (0, -1, 0) & (0, -1, 0) & Yes & Yes \\
7 & (0, -1, 1) & (0, -0.71, 0.7) & No & Yes \\
8 & (0, -1, -1) & (0, -0.71, -0.7) & No & Yes \\
9 & (1, 0, 0) & (1, 0, 0) & Yes & Yes \\
10 & (1, 0, 1) & (0.71, 0, 0.7) & No & Yes \\
11 & (1, 0, -1) & (0.71, 0, -0.7) & No & Yes \\
12 & (1, 1, 0) & (0.71, 0.7, 0) & No & Yes \\
13 & (1, 1, 1) & (1, 1, 1) & Yes & Yes \\
\midrule
14 & (1, 1, -1) & - & No & No \\
\midrule
15 & (1, -1, 0) & (0.71, -0.7, 0) & No & Yes \\
\midrule
16 & (1, -1, 1) & - & No & No \\
\midrule
17 & (1, -1, -1) & (1, -1, -1) & Yes & Yes \\
18 & (-1, 0, 0) & (-1, 0, 0) & Yes & Yes \\
19 & (-1, 0, 1) & (-0.71, 0, 0.7) & No & Yes \\
20 & (-1, 0, -1) & (-0.71, 0, -0.7) & No & Yes \\
21 & (-1, 1, 0) & (-0.71, 0.7, 0) & No & Yes \\
\midrule
22 & (-1, 1, 1) & - & No & No \\
\midrule
23 & (-1, 1, -1) & (-1, 1, -1) & Yes & Yes \\
24 & (-1, -1, 0) & (-0.71, -0.7, 0) & No & Yes \\
25 & (-1, -1, 1) & (-1, -1, 1) & Yes & Yes \\
\midrule
26 & (-1, -1, -1) & - & No & No \\
\bottomrule
\end{tabular*}
\end{table} 

\subsection{Pattern Specific Performance}
Tables~\ref{tab:per-pattern-mae-complete-non-normal-data}-\ref{tab:per-pattern-mae-sparse-downsampled-data} present detailed MAE statistics between target relaxed correlation structures and their Spearman estimations for the complete and sparse non-normal and downsampled data variants.The patterns are ordered by descending MAE to highlight the correlation structures that are the most distorted. In the complete and sparse non-normal variant, all patterns show minimal distortions. Patterns with all positive and negative mixed strong correlations (patterns 13, 17, 23, and 25) are almost perfectly modelled, while patterns that need relaxing (non ideal) become slightly more distorted. For the sparse non-normal variant MAE for non-ideal pattern is minimally higher increasing from $0.04-0.05$. The percentage segments outside of tolerance (OOT) for the complete non-normal variant range from $5.2-10.4\%$ (non ideal patterns), respectively $0\%$ (ideal patterns). For the sparse non-normal variant OOT increases to $24.4-44.4\%$ (non ideal patterns), $0.7-5.7\%$ (simple patterns with no more one strong positive or negative correlation), and remains $0\%$ only for patterns with all strong positive or negative correlations (patterns 13, 17, 23, and 25). For the downsampled variants, pattern 13 $[1, 1, 1]$ has the smallest mean MAE of $0.03-0.04$, while patterns 23 $[-1, 1, -1]$ and 25 $[-1, -1, 1]$ have a mean MAE of $0.21-0.24$ (complete downsampled), respectively $0.15-0.16$ (sparse downsampled). This represents inversion of which patterns are more distorted. For the non-normal variants pattern 23, 25, and 17 are perfectly modelled while but become most error prone in downsampled data. Patterns with no more than a single strong correlation coefficient (patterns 1, 3, 9, 13) keep a low MAE through downsampling ($0.03-0.07$). The MAE for these patterns in the complete variant can move a correlation structure out of their negligible and strong correlation coefficients grouping into the moderate range. Interestingly, sparsification in the downsampled variants slightly improves the accuracy of the correlation structures. To understand why patterns that need relaxation are more likely to fall out of tolerance, we need to recall that the tolerance bands are $\toleranceBands=\{[-1,-0.7],[-0.2,0.2],[0.7,1]\}$ to allow for some distortion but retain clear strong positive, negative, and negligible patterns. MAE is calculated on the basis of the relaxed patterns, as these ensure valid correlation matrices. For interpretation purposes, pattern 23 $[-1, 1, -1]$ (does not need relaxation to be a valid correlation matrix). For this pattern, an MAE of $0.04$ would mean that the empirical correlation might be $[-0.96, 0.96, -0.96]$, which is well within the tolerance band. In comparison, pattern 24 $[-0.71, -0.7, 0]$ needs significant relaxation to become a valid correlation matrix, and the same MAE of $0.04$ would mean that the empirical correlation might be $[-0.68, -0.66, 0.04]$, which is outside the tolerance band. This also means that an MAE of $0.14$ for pattern 24 might result in an empirical correlation structure of $[-0.57, -0.56, 0.14]$. In this case, the previously negative strong correlation coefficients have been distorted into moderate correlation coefficients.

\begin{table}[!ht]
    \caption{MAE statistics per pattern between the target relaxed correlation structure and its Spearman estimate and percentage of segments outside of the tolerance bands $\toleranceBands$ (oot\%) for the \textbf{complete non-normal data variant}. The patterns are ordered by descending MAE.}
    \label{tab:per-pattern-mae-complete-non-normal-data}
    \centering
    \small
    \begin{tabular}{cccccccccccc}
        \toprule
        \multirow{2}{*}{\textbf{ID}} & \textbf{Relaxed} & \multirow{2}{*}{\textbf{Ideal}} & \multirow{2}{*}{\textbf{count}} & \multirow{2}{*}{\textbf{50\%}} & \multirow{2}{*}{\textbf{mean}} & \multirow{2}{*}{\textbf{std}} & \multirow{2}{*}{\textbf{25\%}} & \multirow{2}{*}{\textbf{75\%}} & \multirow{2}{*}{\textbf{min}} & \multirow{2}{*}{\textbf{max}} & \multirow{2}{*}{\textbf{oot\%}} \\
        & \textbf{Structure} & & & & & & & & & &\\
        \midrule
        7 & (0, -0.71, 0.7) & No & 130 & 0.04 & 0.04 & 0.01 & 0.04 & 0.05 & 0.02 & 0.07 & 5.2\\
        15 & (0.71, -0.7, 0) & No & 131 & 0.04 & 0.04 & 0.01 & 0.04 & 0.04 & 0.02 & 0.08 & 10.4\\
        24 & (-0.71, -0.7, 0) & No & 124 & 0.04 & 0.04 & 0.01 & 0.04 & 0.05 & 0.01 & 0.08 & 9.6\\
        21 & (-0.71, 0.7, 0) & No & 132 & 0.04 & 0.04 & 0.01 & 0.04 & 0.04 & 0.02 & 0.07 & 7.4\\
        4 & (0, 0.71, 0.7) & No & 128 & 0.04 & 0.04 & 0.01 & 0.04 & 0.04 & 0.01 & 0.07 & 5.2\\
        5 & (0, 0.71, -0.7) & No & 127 & 0.04 & 0.04 & 0.01 & 0.04 & 0.05 & 0.02 & 0.08 & 5.9\\
        20 & (-0.71, 0, -0.7) & No & 136 & 0.04 & 0.04 & 0.01 & 0.04 & 0.04 & 0.02 & 0.08 & 8.9\\
        8 & (0, -0.71, -0.7) & No & 129 & 0.04 & 0.04 & 0.01 & 0.04 & 0.04 & 0.02 & 0.07 & 6.7\\
        10 & (0.71, 0, 0.7) & No & 131 & 0.04 & 0.04 & 0.01 & 0.04 & 0.04 & 0.02 & 0.09 & 8.9\\
        12 & (0.71, 0.7, 0) & No & 131 & 0.04 & 0.04 & 0.01 & 0.04 & 0.05 & 0.01 & 0.09 & 7.4\\
        11 & (0.71, 0, -0.7) & No & 131 & 0.04 & 0.04 & 0.01 & 0.03 & 0.04 & 0.02 & 0.07 & 9.6\\
        19 & (-0.71, 0, 0.7) & No & 132 & 0.04 & 0.04 & 0.01 & 0.03 & 0.04 & 0.02 & 0.06 & 8.9\\
        18 & (-1, 0, 0) & Yes & 130 & 0.01 & 0.01 & 0.01 & 0.00 & 0.01 & 0.00 & 0.06 & 0\\
        0 & (0, 0, 0) & Yes & 132 & 0.01 & 0.01 & 0.01 & 0.00 & 0.01 & 0.00 & 0.05 & 0\\
        1 & (0, 0, 1) & Yes & 129 & 0.01 & 0.01 & 0.01 & 0.00 & 0.01 & 0.00 & 0.05 & 0\\
        9 & (1, 0, 0) & Yes & 131 & 0.01 & 0.01 & 0.01 & 0.00 & 0.01 & 0.00 & 0.04 & 0\\
        6 & (0, -1, 0) & Yes & 135 & 0.01 & 0.01 & 0.01 & 0.00 & 0.01 & 0.00 & 0.04 & 0\\
        2 & (0, 0, -1) & Yes & 130 & 0.01 & 0.01 & 0.01 & 0.00 & 0.01 & 0.00 & 0.05 & 0\\
        3 & (0, 1, 0) & Yes & 129 & 0.00 & 0.01 & 0.01 & 0.00 & 0.01 & 0.00 & 0.06 & 0\\
        13 & (1, 1, 1) & Yes & 130 & 0.00 & 0.00 & 0.01 & 0.00 & 0.00 & 0.00 & 0.03 & 0\\
        17 & (1, -1, -1) & Yes & 132 & 0.00 & 0.00 & 0.01 & 0.00 & 0.00 & 0.00 & 0.03 & 0\\
        23 & (-1, 1, -1) & Yes & 129 & 0.00 & 0.00 & 0.01 & 0.00 & 0.00 & 0.00 & 0.03 & 0\\
        25 & (-1, -1, 1) & Yes & 131 & 0.00 & 0.00 & 0.01 & 0.00 & 0.00 & 0.00 & 0.03 & 0\\
        \bottomrule
    \end{tabular}
\end{table}

\begin{table}[!ht]
    \caption{MAE statistics per pattern between the target relaxed correlation structure and its Spearman estimate and percentage of segments outside of the tolerance bands $\toleranceBands$ (oot\%) for the \textbf{complete downsampled data variant}. The patterns are ordered by descending MAE.}
    \label{tab:per-pattern-mae-complete-downsampled-data}
    \centering
    \small
    \begin{tabular}{cccccccccccc}
        \toprule
        \multirow{2}{*}{\textbf{ID}} & \textbf{Relaxed} & \multirow{2}{*}{\textbf{Ideal}} & \multirow{2}{*}{\textbf{count}} & \multirow{2}{*}{\textbf{50\%}} & \multirow{2}{*}{\textbf{mean}} & \multirow{2}{*}{\textbf{std}} & \multirow{2}{*}{\textbf{25\%}} & \multirow{2}{*}{\textbf{75\%}} & \multirow{2}{*}{\textbf{min}} & \multirow{2}{*}{\textbf{max}} & \multirow{2}{*}{\textbf{oot\%}} \\
        & \textbf{Structure} & & & & & & & & & &\\
        \midrule
        23 & (-1, 1, -1) & Yes & 129 & 0.23 & 0.24 & 0.08 & 0.18 & 0.28 & 0.06 & 0.61 & 69.6\\
        25 & (-1, -1, 1) & Yes & 131 & 0.21 & 0.21 & 0.09 & 0.14 & 0.27 & 0.03 & 0.52 & 62.2\\
        24 & (-0.71, -0.7, 0) & No & 124 & 0.18 & 0.19 & 0.07 & 0.14 & 0.23 & 0.08 & 0.43 & 91.8\\
        18 & (-1, 0, 0) & Yes & 130 & 0.18 & 0.20 & 0.08 & 0.14 & 0.23 & 0.06 & 0.48 & 73.3\\
        17 & (1, -1, -1) & Yes & 132 & 0.17 & 0.18 & 0.06 & 0.13 & 0.20 & 0.08 & 0.50 & 28.9\\
        21 & (-0.71, 0.7, 0) & No & 132 & 0.16 & 0.17 & 0.07 & 0.12 & 0.20 & 0.07 & 0.53 & 97.8\\
        20 & (-0.71, 0, -0.7) & No & 136 & 0.16 & 0.18 & 0.06 & 0.14 & 0.19 & 0.10 & 0.41 & 100\\
        19 & (-0.71, 0, 0.7) & No & 132 & 0.13 & 0.14 & 0.05 & 0.11 & 0.16 & 0.07 & 0.34 & 97.8\\
        8 & (0, -0.71, -0.7) & No & 129 & 0.13 & 0.15 & 0.07 & 0.10 & 0.16 & 0.08 & 0.55 & 95.6\\
        6 & (0, -1, 0) & Yes & 135 & 0.13 & 0.14 & 0.07 & 0.09 & 0.17 & 0.04 & 0.41 & 40.7\\
        15 & (0.71, -0.7, 0) & No & 131 & 0.12 & 0.13 & 0.07 & 0.07 & 0.15 & 0.04 & 0.47 & 97\\
        2 & (0, 0, -1) & Yes & 130 & 0.12 & 0.14 & 0.06 & 0.10 & 0.16 & 0.06 & 0.38 & 19.3\\
        7 & (0, -0.71, 0.7) & No & 130 & 0.10 & 0.11 & 0.06 & 0.08 & 0.13 & 0.04 & 0.41 & 96.3\\
        11 & (0.71, 0, -0.7) & No & 131 & 0.10 & 0.12 & 0.07 & 0.09 & 0.12 & 0.06 & 0.50 & 97\\
        5 & (0, 0.71, -0.7) & No & 127 & 0.10 & 0.12 & 0.06 & 0.08 & 0.13 & 0.06 & 0.36 & 94.1\\
        12 & (0.71, 0.7, 0) & No & 131 & 0.08 & 0.09 & 0.06 & 0.05 & 0.11 & 0.02 & 0.34 & 93.3\\
        3 & (0, 1, 0) & Yes & 129 & 0.07 & 0.08 & 0.07 & 0.04 & 0.11 & 0.01 & 0.34 & 14.8\\
        10 & (0.71, 0, 0.7) & No & 131 & 0.06 & 0.09 & 0.06 & 0.05 & 0.12 & 0.04 & 0.38 & 95.6\\
        4 & (0, 0.71, 0.7) & No & 128 & 0.06 & 0.08 & 0.05 & 0.05 & 0.09 & 0.03 & 0.34 & 90.4\\
        0 & (0, 0, 0) & Yes & 132 & 0.06 & 0.09 & 0.09 & 0.04 & 0.13 & 0.00 & 0.47 & 25.2\\
        1 & (0, 0, 1) & Yes & 129 & 0.05 & 0.07 & 0.06 & 0.03 & 0.08 & 0.01 & 0.30 & 11.8\\
        9 & (1, 0, 0) & Yes & 131 & 0.05 & 0.07 & 0.06 & 0.03 & 0.08 & 0.00 & 0.42 & 8.9\\
        13 & (1, 1, 1) & Yes & 130 & 0.03 & 0.04 & 0.03 & 0.02 & 0.06 & 0.01 & 0.16 & 0\\
        \bottomrule
    \end{tabular}
\end{table}

\begin{table}[!ht]
    \caption{MAE statistics per pattern between the target relaxed correlation structure and its Spearman estimate and percentage of segments outside of the tolerance bands $\toleranceBands$ (oot\%) for the \textbf{sparse non-normal data variant}. The patterns are ordered by descending MAE.}
    \label{tab:per-pattern-mae-sparse-non-normal-data}
    \centering
    \small
    \begin{tabular}{cccccccccccc}
        \toprule
        \multirow{2}{*}{\textbf{ID}} & \textbf{Relaxed} & \multirow{2}{*}{\textbf{Ideal}} & \multirow{2}{*}{\textbf{count}} & \multirow{2}{*}{\textbf{50\%}} & \multirow{2}{*}{\textbf{mean}} & \multirow{2}{*}{\textbf{std}} & \multirow{2}{*}{\textbf{25\%}} & \multirow{2}{*}{\textbf{75\%}} & \multirow{2}{*}{\textbf{min}} & \multirow{2}{*}{\textbf{max}} & \multirow{2}{*}{\textbf{oot\%}} \\
        & \textbf{Structure} & & & & & & & & & &\\
        \midrule
        24 & (-0.71, -0.7, 0) & No & 124 & 0.05 & 0.05 & 0.02 & 0.04 & 0.06 & 0.01 & 0.11 & 24.4\\
        4 & (0, 0.71, 0.7) & No & 128 & 0.05 & 0.05 & 0.02 & 0.04 & 0.06 & 0.01 & 0.15 & 24.4\\
        5 & (0, 0.71, -0.7) & No & 127 & 0.04 & 0.05 & 0.02 & 0.03 & 0.06 & 0.01 & 0.14 & 25.2\\
        8 & (0, -0.71, -0.7) & No & 129 & 0.04 & 0.05 & 0.02 & 0.04 & 0.06 & 0.01 & 0.14 & 31.8\\
        15 & (0.71, -0.7, 0) & No & 131 & 0.04 & 0.05 & 0.03 & 0.03 & 0.06 & 0.01 & 0.15 & 31.8\\
        11 & (0.71, 0, -0.7) & No & 131 & 0.04 & 0.05 & 0.03 & 0.03 & 0.06 & 0.01 & 0.17 & 44.4\\
        12 & (0.71, 0.7, 0) & No & 131 & 0.04 & 0.05 & 0.02 & 0.04 & 0.05 & 0.01 & 0.14 & 31.1\\
        20 & (-0.71, 0, -0.7) & No & 136 & 0.04 & 0.05 & 0.02 & 0.03 & 0.05 & 0.01 & 0.12 & 34.1\\
        19 & (-0.71, 0, 0.7) & No & 132 & 0.04 & 0.05 & 0.02 & 0.03 & 0.06 & 0.01 & 0.12 & 37\\
        21 & (-0.71, 0.7, 0) & No & 132 & 0.04 & 0.05 & 0.03 & 0.03 & 0.05 & 0.01 & 0.15 & 37.8\\
        7 & (0, -0.71, 0.7) & No & 130 & 0.04 & 0.05 & 0.02 & 0.03 & 0.06 & 0.02 & 0.14 & 27.4\\
        10 & (0.71, 0, 0.7) & No & 131 & 0.04 & 0.05 & 0.02 & 0.03 & 0.06 & 0.01 & 0.16 & 39.3\\
        0 & (0, 0, 0) & Yes & 132 & 0.03 & 0.04 & 0.03 & 0.02 & 0.05 & 0.00 & 0.18 & 5.2\\
        1 & (0, 0, 1) & Yes & 129 & 0.02 & 0.03 & 0.03 & 0.01 & 0.04 & 0.00 & 0.16 & 0.7\\
        18 & (-1, 0, 0) & Yes & 130 & 0.02 & 0.03 & 0.03 & 0.01 & 0.03 & 0.00 & 0.16 & 1.5\\
        9 & (1, 0, 0) & Yes & 131 & 0.02 & 0.03 & 0.03 & 0.01 & 0.03 & 0.00 & 0.17 & 1.5\\
        6 & (0, -1, 0) & Yes & 135 & 0.02 & 0.03 & 0.03 & 0.01 & 0.03 & 0.00 & 0.16 & 1.5\\
        3 & (0, 1, 0) & Yes & 129 & 0.01 & 0.02 & 0.03 & 0.01 & 0.03 & 0.00 & 0.19 & 0.7\\
        2 & (0, 0, -1) & Yes & 130 & 0.01 & 0.03 & 0.03 & 0.01 & 0.03 & 0.00 & 0.18 & 1.5\\
        13 & (1, 1, 1) & Yes & 130 & 0.00 & 0.00 & 0.01 & 0.00 & 0.00 & 0.00 & 0.03 & 0\\
        17 & (1, -1, -1) & Yes & 132 & 0.00 & 0.00 & 0.01 & 0.00 & 0.00 & 0.00 & 0.03 & 0\\
        23 & (-1, 1, -1)  & Yes & 129 & 0.00 & 0.00 & 0.01 & 0.00 & 0.00 & 0.00 & 0.03 & 0\\
        25 & (-1, -1, 1) & Yes & 131 & 0.00 & 0.00 & 0.01 & 0.00 & 0.00 & 0.00 & 0.03 & 0\\
        \bottomrule
    \end{tabular}
\end{table}

\begin{table}[!ht]
    \caption{MAE statistics per pattern between the target relaxed correlation structure and its Spearman estimate and percentage of segments outside of the tolerance bands $\toleranceBands$ (oot\%) for the \textbf{sparse downsampled data variant}. The patterns are ordered by descending MAE.}
    \label{tab:per-pattern-mae-sparse-downsampled-data}
    \centering
    \small
     \begin{tabular}{cccccccccccc}
        \toprule
        \multirow{2}{*}{\textbf{ID}} & \textbf{Relaxed} & \multirow{2}{*}{\textbf{Ideal}} & \multirow{2}{*}{\textbf{count}} & \multirow{2}{*}{\textbf{50\%}} & \multirow{2}{*}{\textbf{mean}} & \multirow{2}{*}{\textbf{std}} & \multirow{2}{*}{\textbf{25\%}} & \multirow{2}{*}{\textbf{75\%}} & \multirow{2}{*}{\textbf{min}} & \multirow{2}{*}{\textbf{max}} & \multirow{2}{*}{\textbf{oot\%}} \\
        & \textbf{Structure} & & & & & & & & & &\\
        \midrule
        23 & (-1, 1, -1) & Yes & 129 & 0.16 & 0.18 & 0.07 & 0.14 & 0.22 & 0.05 & 0.42 & 37.8\\
        25 & (-1, -1, 1) & Yes & 131 & 0.15 & 0.16 & 0.05 & 0.12 & 0.19 & 0.07 & 0.36 & 40\\
        18 & (-1, 0, 0) & Yes & 130 & 0.15 & 0.16 & 0.06 & 0.12 & 0.19 & 0.04 & 0.40 & 51.1\\
        24 & (-0.71, -0.7, 0) & No & 124 & 0.14 & 0.15 & 0.06 & 0.11 & 0.18 & 0.05 & 0.40 & 91.8\\
        20 & (-0.71, 0, -0.7) & No & 136 & 0.13 & 0.14 & 0.05 & 0.11 & 0.15 & 0.08 & 0.43 & 100\\
        17 & (1, -1, -1) & Yes & 132 & 0.12 & 0.13 & 0.04 & 0.10 & 0.15 & 0.05 & 0.28 & 8.9\\
        21 & (-0.71, 0.7, 0) & No & 132 & 0.11 & 0.12 & 0.05 & 0.09 & 0.15 & 0.05 & 0.36 & 97.8\\
        8 & (0, -0.71, -0.7) & No & 129 & 0.11 & 0.12 & 0.05 & 0.09 & 0.13 & 0.04 & 0.39 & 95.6\\
        2 & (0, 0, -1) & Yes & 130 & 0.11 & 0.13 & 0.07 & 0.08 & 0.15 & 0.05 & 0.41 & 17\\
        19 & (-0.71, 0, 0.7) & No & 132 & 0.10 & 0.12 & 0.05 & 0.09 & 0.13 & 0.06 & 0.30 & 97.8\\
        6 & (0, -1, 0) & Yes & 135 & 0.10 & 0.11 & 0.06 & 0.07 & 0.14 & 0.02 & 0.37 & 19.3\\
        15 & (0.71, -0.7, 0) & No & 131 & 0.09 & 0.10 & 0.05 & 0.06 & 0.12 & 0.04 & 0.30 & 94.8\\
        5 & (0, 0.71, -0.7) & No & 127 & 0.08 & 0.11 & 0.06 & 0.07 & 0.12 & 0.04 & 0.32 & 94.1\\
        11 & (0.71, 0, -0.7) & No & 131 & 0.08 & 0.11 & 0.06 & 0.07 & 0.12 & 0.04 & 0.34 & 97\\
        7 & (0, -0.71, 0.7) & No & 130 & 0.08 & 0.10 & 0.07 & 0.06 & 0.10 & 0.04 & 0.52 & 96.3\\
        10 & (0.71, 0, 0.7) & No & 131 & 0.06 & 0.08 & 0.05 & 0.04 & 0.09 & 0.02 & 0.28 & 92.6\\
        12 & (0.71, 0.7, 0) & No & 131 & 0.06 & 0.08 & 0.07 & 0.05 & 0.09 & 0.03 & 0.39 & 93.3\\
        1 & (0, 0, 1) & Yes & 129 & 0.06 & 0.07 & 0.06 & 0.03 & 0.08 & 0.01 & 0.32 & 12.6\\
        4 & (0, 0.71, 0.7) & No & 128 & 0.05 & 0.07 & 0.05 & 0.04 & 0.09 & 0.02 & 0.37 & 87.4\\
        0 & (0, 0, 0) & Yes & 132 & 0.05 & 0.09 & 0.08 & 0.04 & 0.13 & 0.01 & 0.38 & 23.7\\
        9 & (1, 0, 0) & Yes & 131 & 0.05 & 0.07 & 0.06 & 0.02 & 0.08 & 0.003 & 0.33 & 11.1\\
        3 & 0, 1, 0) & Yes & 129 & 0.04 & 0.07 & 0.07 & 0.02 & 0.08 & 0.001 & 0.40 & 11.1\\
        13 & (1, 1, 1) & Yes & 130 & 0.02 & 0.03 & 0.02 & 0.02 & 0.03 & 0.004 & 0.15 & 0\\
        \bottomrule
    \end{tabular}
\end{table}

\subsection{Aggregated Correlation Structures}
\begin{figure}[h]
    \centering
    \begin{tabular}{cc}
        \includegraphics[width=0.48\textwidth, height=0.3\textheight, keepaspectratio]{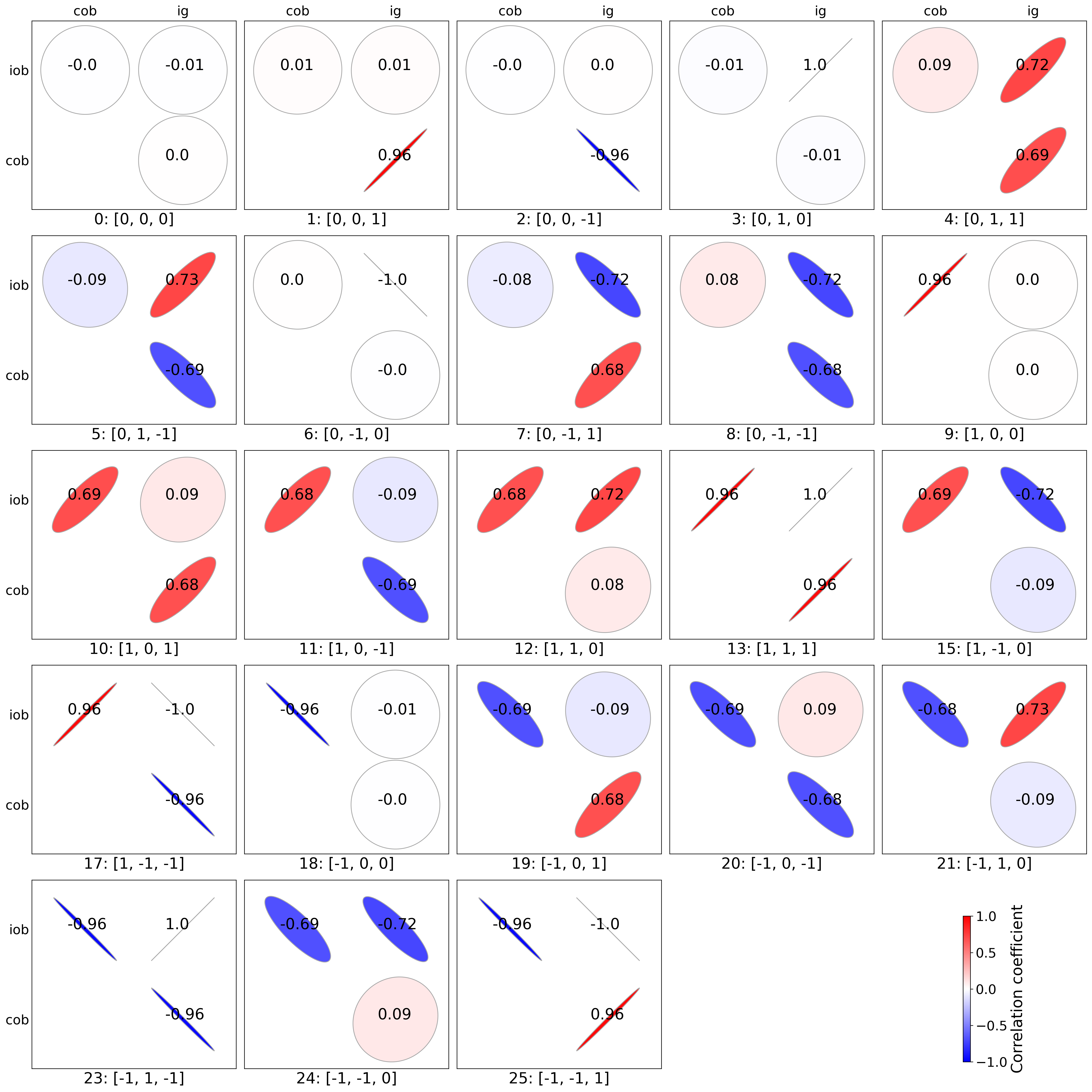} &
        \includegraphics[width=0.48\textwidth, height=0.3\textheight, keepaspectratio]{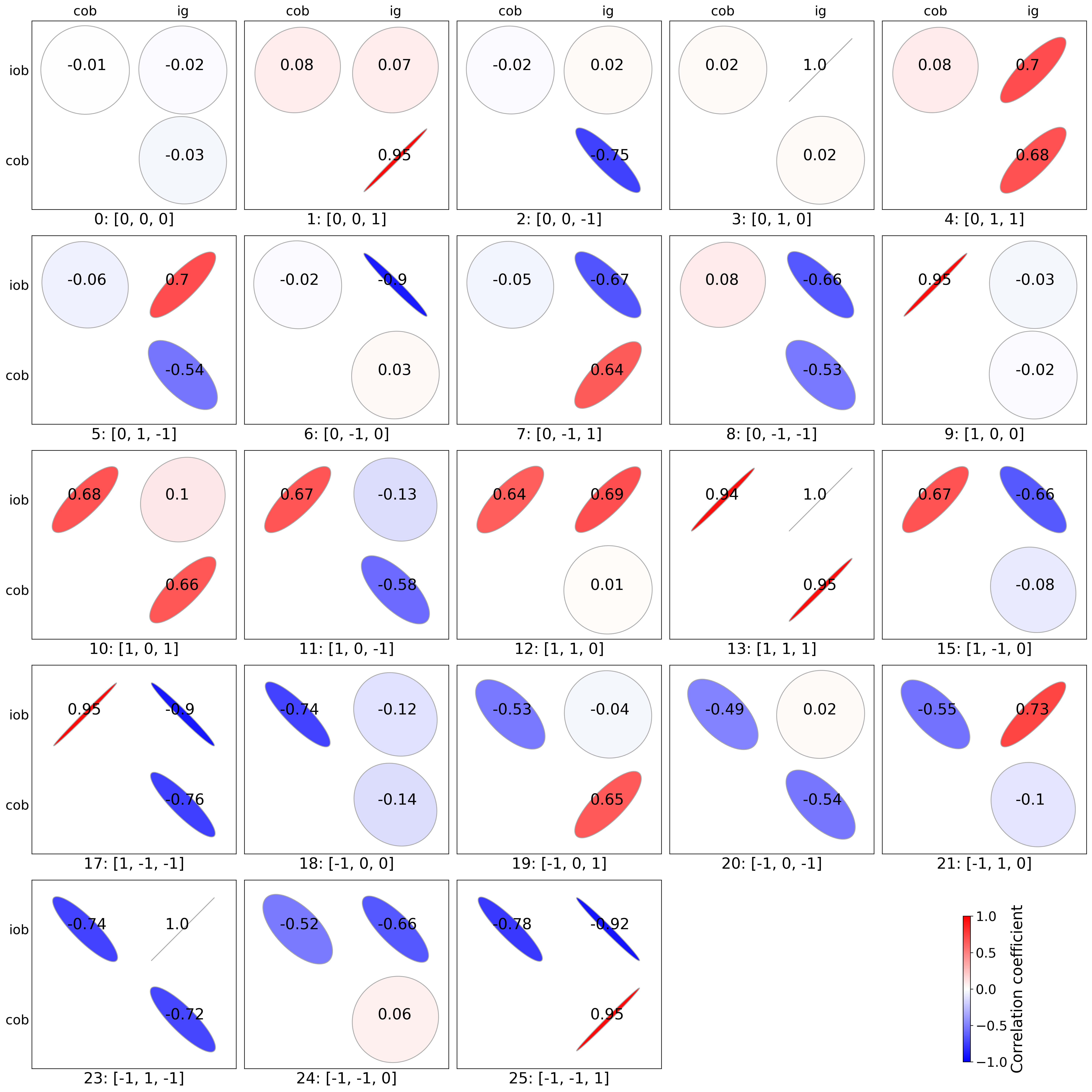}\\
        (a) Cor structures: non-normal, partial & (b) Cor structures: downsampled, partial \\
    \end{tabular}
    \caption{Correlation structures visualisation of Spearman estimates calculated using aggregated observations from all segments of a pattern shows that downsampling degrades correlation structures. (a) Non-normal data with $6.03\%$ segments outside tolerance bands, mean MAE$=0.03$, and mean segment length=$8848$ observations (min $592$). (b) Downsampled data with $67.2\%$ segments outside tolerance bands, mean MAE$=0.13$, and mean segment length=$211$ observations (min $15$). Data for exploratory subject trim-fire-24 for the partial data variants.}
    \label{fig:correlations-partial-nn-downsampled}
\end{figure}

\begin{figure}[h]
    \centering
    \begin{tabular}{cc}
        \includegraphics[width=0.48\textwidth, height=0.3\textheight, keepaspectratio]{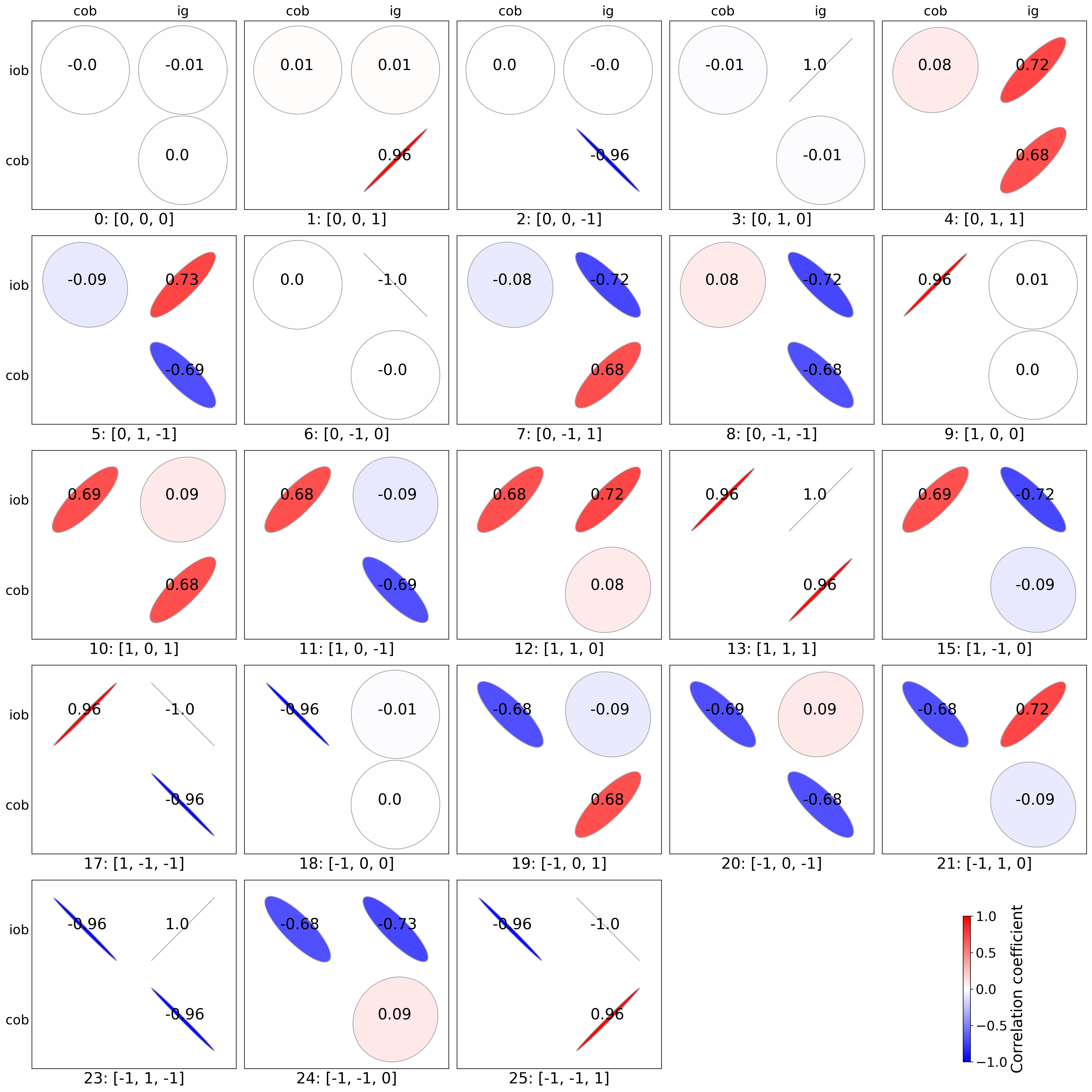} &
        \includegraphics[width=0.48\textwidth, height=0.3\textheight, keepaspectratio]{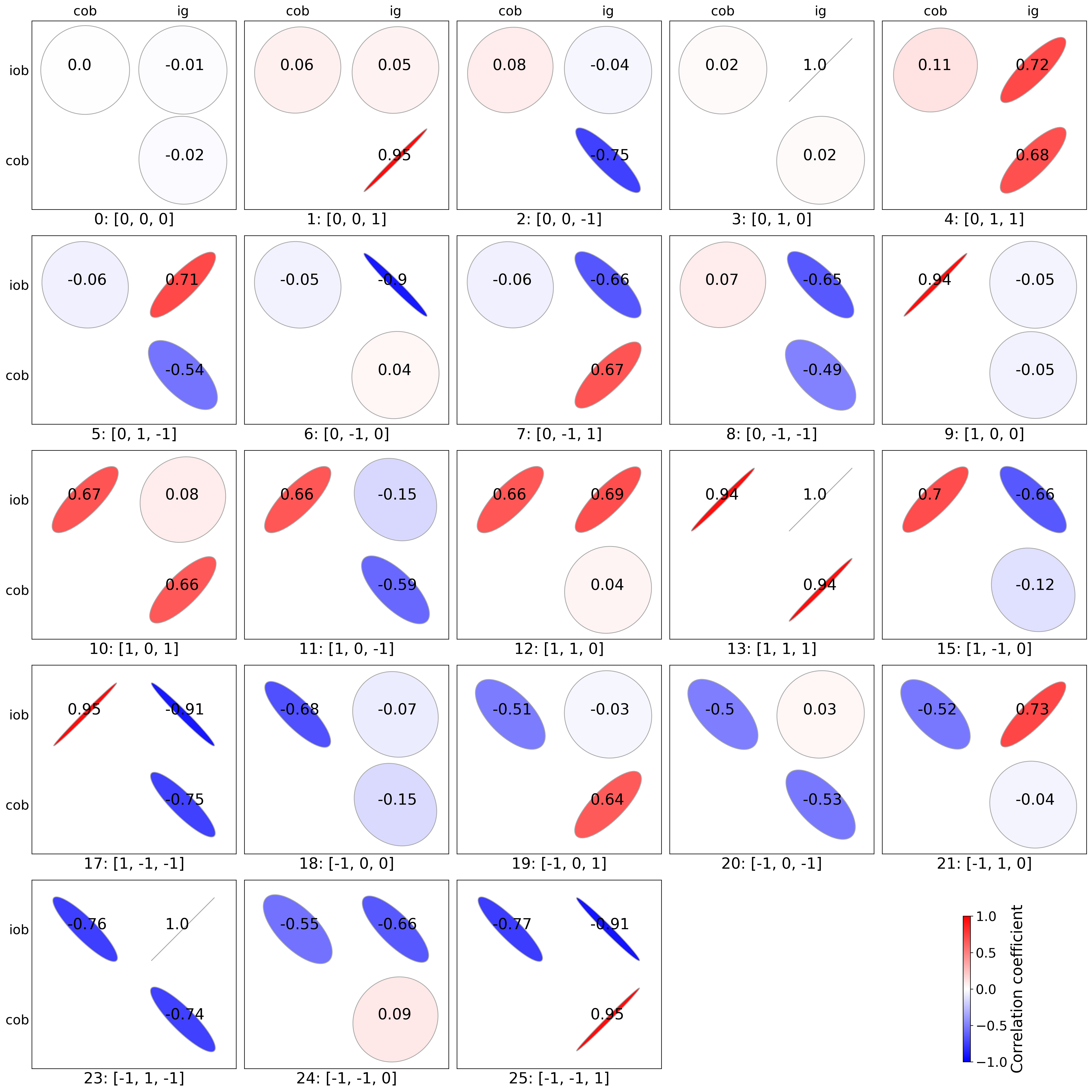}\\
        (a) Cor structures: non-normal, complete & (b) Cor structures: downsampled, complete \\
    \end{tabular}
    \caption{Correlation structures visualisation of Spearman estimates calculated using aggregated observations from all segments of a pattern shows that downsampling degrades correlation structures. (a) Non-normal data with $4.23\%$ segments outside tolerance bands, mean MAE$=0.02$, and mean segment length=$12640$ observations (min $900$). (b) Downsampled data with $67.6\%$ segments outside tolerance bands, mean MAE$=0.13$, and mean segment length=$211$ observations (min $15$). Data for exploratory subject trim-fire-24 for the complete variants.}
    \label{fig:correlations-complete-nn-downsampled}
\end{figure}

Figures~\ref{fig:correlations-partial-nn-downsampled} and \ref{fig:correlations-complete-nn-downsampled} illustrate that downsampling degrades correlation structures shown as ellipse plots, where the orientation and shape of each ellipse represent the direction and strength of correlations between time series variates. In the non-normal complete variant (subplots a), the correlation patterns are crisper and more distinct, while in the downsampled complete data (subplots b), the same patterns are still discernible but with reduced definition. Complex patterns with multiple negative correlations such as patterns 23 $[-1, 1, -1]$ and 25 $[-1,-1,1]$ are affected more strongly (mean MAE: $0.16-0.24$) than pattern 13 $[1, 1, 1]$ (mean MAE: $0.02-0.03$) that shows minimal distortions. Both figures use data from subject trim-fire-24 in the exploratory dataset and clearly visualise why the statistical measures show an increase in MAE and more segments outside tolerance bands after downsampling, particularly for irregular data.

\clearpage

\section{Correlation Measures}\label{app:correlation-measures}

\subsection{Impact of Segment length}
Table~\ref{tab:minimal-segment-lengths-mae-stats} quantifies the relationship between segment length and Spearman correlation estimation accuracy for the complete non-normal data variant. MAE values decrease steadily as the length of the segments increases, with a notable threshold around $30$ observations (MAE$= 0.090$), below which the estimation of the correlation structure becomes increasingly unreliable, contributing errors $>0.1$ to MAE. For $60$ observations, 75\% of segments achieve MAE$<0.1$. Higher correlation estimation accuracy allows researchers to exclude estimation errors as a reason for correlation pattern deformation. From these findings, we recommend that researchers work with high-frequency data for correlation-based clustering to increase the chances of reasonable segment lengths.

\begin{table}[!ht]
    \caption{MAE statistics between specified and Spearman correlation estimations for different segment lengths calculated across the $30$ subjects of the \textbf{complete non-normal data variant}.}
    \label{tab:minimal-segment-lengths-mae-stats}
    \centering
    \begin{tabular}{ccccc}
        \toprule
        \textbf{Length} & \textbf{Mean} & \textbf{Median} & \textbf{25\%} & \textbf{75\%} \\
        \midrule
        10 & 0.161 & 0.145 & 0.057 & 0.237 \\
        15 & 0.128 & 0.117 & 0.047 & 0.192 \\
        20 & 0.109 & 0.101 & 0.039 & 0.162 \\
        30 & 0.090 & 0.082 & 0.032 & 0.131 \\
        60 & 0.065 & 0.059 & 0.025 & 0.096 \\
        80 & 0.058 & 0.053 & 0.022 & 0.087 \\
        100 & 0.053 & 0.049 & 0.020 & 0.078 \\
        200 & 0.042 & 0.038 & 0.015 & 0.061 \\
        400 & 0.034 & 0.032 & 0.013 & 0.051 \\
        600 & 0.031 & 0.030 & 0.012 & 0.047 \\
        800 & 0.029 & 0.029 & 0.011 & 0.044 \\
        \bottomrule
    \end{tabular}
\end{table}

\subsection{Correlation Measure Comparison}
Tables~\ref{tab:overall_mae_per_cor_per_data_variant} and \ref{tab:seg_out_tolerance_per_cor_per_data_variant} compare the performance of Spearman, Pearson, and Kendall correlation measures across all data variants. Table~\ref{tab:overall_mae_per_cor_per_data_variant} presents the overall MAE values, while Table \ref{tab:seg_out_tolerance_per_cor_per_data_variant} shows the number of segments with correlation estimates outside the tolerance bands. Spearman correlation consistently outperforms alternatives, particularly for non-normal data variants where Pearson's MAE increases significantly. Kendall's correlation performs poorly across all variants, with substantially more segments falling outside the tolerance bands than either Spearman or Pearson.

\begin{table}[!ht]
\caption{Overall MAE for different correlation measures and different data variants}
\label{tab:overall_mae_per_cor_per_data_variant}
\small
\setlength{\tabcolsep}{4pt}
\centering
\begin{tabular}{lrrrrrrrrrrrrr}
\toprule
& \multicolumn{4}{c}{\textbf{Spearman}} & \multicolumn{4}{c}{\textbf{Pearson}} & \multicolumn{4}{c}{\textbf{Kendal}} \\
\cmidrule(lr){2-5} \cmidrule(lr){6-9} \cmidrule(lr){10-13}
\textbf{Data Variant} & \textbf{mean} & \textbf{50\%} & \textbf{25\%} & \textbf{75\%} & \textbf{mean} & \textbf{50\%} & \textbf{25\%} & \textbf{75\%} & \textbf{mean} & \textbf{50\%} & \textbf{25\%} & \textbf{75\%} \\
\midrule
\textbf{Raw} &  & &  & &  & &  &  & & &  & \\
100\% & 0.51 & 0.47 & 0.35 & 0.48 & 0.51 & 0.47 & 0.35 & 0.48 & 0.51 & 0.47 & 0.34 & 0.48 \\
70\% & 0.51 & 0.47 & 0.35 & 0.48 & 0.51 & 0.47 & 0.35 & 0.48 & 0.51 & 0.47 & 0.34 & 0.48 \\
10\% & 0.52 & 0.47 & 0.37 & 0.50 & 0.52 & 0.47 & 0.37 & 0.50 & 0.52 & 0.47 & 0.36 & 0.49 \\
\midrule
\textbf{Correlated} &  & &  & &  & &  &  & & &  & \\
100\% & 0.02 & 0.03 & 0.002 & 0.04 & 0.03 & 0.04 & 0.002 & 0.05 & 0.07 & 0.14 & 0.001 & 0.14 \\
70\% & 0.03 & 0.03 & 0.002 & 0.04 & 0.03 & 0.04 & 0.003 & 0.05 & 0.07 & 0.14 & 0.002 & 0.14 \\
10\% & 0.03 & 0.03 & 0.01 & 0.05 & 0.04 & 0.04 & 0.01 & 0.06 & 0.08 & 0.14 & 0.01 & 0.14 \\
\midrule
\textbf{Non-normal} &  & &  & &  & &  &  & & &  & \\
100\% & 0.02 & 0.03 & 0.004 & 0.04 & 0.10 & 0.08 & 0.04 & 0.14 & 0.08 & 0.13 & 0.01 & 0.13 \\
70\% & 0.03 & 0.03 & 0.01 & 0.04 & 0.10 & 0.08 & 0.05 & 0.14 & 0.08 & 0.13 & 0.01 & 0.13 \\
10\% & 0.03 & 0.03 & 0.01 & 0.05 & 0.10 & 0.09 & 0.05 & 0.14 & 0.08 & 0.13 & 0.02 & 0.13 \\
\midrule
\textbf{Downsampled} &  & &  & &  & &  &  & & &  & \\
100\% & 0.13 & 0.12 & 0.07 & 0.17 & 0.13 & 0.12 & 0.07 & 0.17 & 0.21 & 0.2 & 0.16 & 0.25 \\
70\% & 0.13 & 0.11 & 0.07 & 0.17 & 0.13 & 0.11 & 0.07 & 0.17 & 0.20 & 0.20 & 0.16 & 0.25 \\
10\% & 0.11 & 0.10 & 0.06 & 0.14 & 0.13 & 0.11 & 0.07 & 0.17 & 0.18 & 0.18 & 0.14 & 0.22 \\
\bottomrule
\end{tabular}
\end{table}

\begin{table}[!ht]
\caption{Segments outside tolerance for different correlation measures and the different data variants}
\label{tab:seg_out_tolerance_per_cor_per_data_variant}
\centering
\small
\setlength{\tabcolsep}{4pt}
\begin{tabular}{lrrrrrrrrrrrrr}
\toprule
& \multicolumn{4}{c}{\textbf{Spearman}} & \multicolumn{4}{c}{\textbf{Pearson}} & \multicolumn{4}{c}{\textbf{Kendal}} \\
\cmidrule(lr){2-5} \cmidrule(lr){6-9} \cmidrule(lr){10-13}
\textbf{Data Variant} & \textbf{mean} & \textbf{50\%} & \textbf{25\%} & \textbf{75\%} & \textbf{mean} & \textbf{50\%} & \textbf{25\%} & \textbf{75\%} & \textbf{mean} & \textbf{50\%} & \textbf{25\%} & \textbf{75\%} \\
\midrule
\textbf{Raw} &  & &  & &  & &  &  & & &  & \\
100\% & 95.6 & 95 & 96 & 96 & 95.6 & 95 & 96 & 96 & 95.6 & 95 & 96 & 96 \\
70\% & 95.6 & 95 & 96 & 96 & 95.6 & 95 & 96 & 96 & 95.6 & 95 & 96 & 96 \\
10\% & 95.8 & 95 & 96 & 96 & 95.9 & 95 & 96 & 96 & 95.6 & 95 & 96 & 96 \\
\midrule
\textbf{Correlated} &  & &  & &  & &  &  & & &  & \\
100\% & 1.9 & 1 & 2 & 2.8 & 0.1 & 0 & 0 & 0 & 52.1 & 51 & 52 & 53 \\
70\% & 3.1 & 2 & 3 & 4 & 0.4 & 0 & 0 & 1 & 52.1 & 51 & 52 & 53 \\
10\% & 14.6 & 13 & 14.5 & 16 & 7.6 & 6 & 8 & 9 & 52.1 & 51 & 52 & 53 \\
\midrule
\textbf{Non-normal} &  & &  & &  & &  &  & & &  & \\
100\% & 4.2 & 1 & 2.5 & 4 & 61.8 & 49.3 & 65 & 70.8 & 52.1 & 51 & 52 & 53 \\
70\% & 6 & 3 & 4 & 6 & 61.9 & 50 & 64.5 & 70.8 & 52.1 & 51 & 52 & 53 \\
10\% & 18.1 & 15 & 17 & 19 & 61.7 & 52 & 64 & 69.8 & 52.1 & 51 & 52 & 53 \\
\midrule
\textbf{Downsampled} &  & &  & &  & &  &  & & &  & \\
100\% & 67.6 & 62 & 68.5 & 73 & 66 & 58 & 67.5 & 73 & 79.7 & 79 & 79.5 & 81 \\
70\% & 67.2 & 67 & 61 & 72.8 & 66.4 & 67 & 56.5 & 74.8 & 79.2 & 79 & 77.3 & 81 \\
10\% & 61.7 & 61 & 58 & 64.8 & 66 & 66.5 & 59 & 71.5 & 77.6 & 78 & 76 & 79 \\
\bottomrule
\end{tabular}
\end{table}

\clearpage

\section{Benchmark Reference Values}\label{app:benchmark-reference-values}
\subsection{Purpose and Overview}
The benchmark reference values in this appendix serve as calibration points for evaluating correlation-based clustering algorithms. \dbname{}'s synthetic nature provides precisely defined correlation structures and allows analysis of how distribution shifts, sparsification, and downsampling affect these structures in isolation, hence creating unambiguous ground truth unavailable in real-world datasets.

These reference values enable researchers to quantitatively assess the performance of the algorithm in relation data variation and clustering mistakes. Although internal clustering quality indices (such as SWC, DBI) measure how well an algorithm groups similar objects (optimizing for within cluster similarity and between cluster separation), without reference values for correlation structures, interpreting these metrics is challenging. Our controlled degradation conditions allow researchers to contextualise an algorithms' mistakes by comparing them to known segmentation and clustering errors, facilitating both results interpretation and objective hyperparameter optimisation.

The tables that follow map specific error conditions (segmentation errors, misclassified segments) to the resulting performance measures (SWC, DBI, Jaccard, MAE), establishing the foundation for the standardised evaluation protocol detailed in Section~\ref{sec:benchmark-usage}.

\subsection{Performance Measures}
Using \dbname{}, we have extensively evaluated various distance measures and internal indices to determine their suitability for comparing correlation matrices and assess within-cluster similarity and between-cluster separation for correlation structures. The results of this research showed that the internal validity indices SWC and DBI perform better than the Calinski-Harabasz Index (VRC) and the Pakhira-Bandyopadhyay-Maulik Index (PBM) using the L5 norm as distance measure between correlation structures. For the classification of correlation structures into their target pattern, the L1 norm distance measure performed better than other Lp norms and more sophisticated distance measures such as the Förstern metric \cite{Foerstner2003} or the Log Frobenius distance \cite{Ergezer2018}. Here we give the formulas for calculating these measures. The formula for calculating MAE has been defined in Section~\ref{sec:benchmark-validation}. Please refer to our GitHub repository \url{https://github.com/isabelladegen/corrclust-validation} for the implementation of these measures.

\paragraph{Lp-norm Distance}
The Lp norm distance (also called Minkowski) between any two correlation matrices is: 
\begin{equation}
        \distanceMeasure[L_p] = \left(\sum_{i=1}^{\nVariates}\sum_{j>i}^{\nVariates}(\correlationMatrixElement_{ij}-\canonicalPatternElement_{ij})^p\right)^\frac{1}{p},
        \label{eq:lp_norm}
\end{equation}
where $\correlationMatrix$ is the empirical correlation matrix of segment $\segIndex$ and $\relaxedPattern$ is the correlation matrix of the target pattern $\patternIndex$. Note that we only used the upper half of the correlation matrix, making this a vector distance between the correlation coefficients. We use $p=1$ to map the correlation structures ($\correlationMatrix$) to ground truth ($\relaxedPattern$) and $p=5$ to calculate the internal indices.

\paragraph{SWC - Silhouette Width Criterion}
The silhouette width criterion (SWC) \cite{Rousseeuw1987} is a measure bounded by $[-1,1]$ that evaluates how similar an object (in our case, a segment's correlation matrix $\correlationMatrix$) is to other segment's correlation structure in its own cluster compared to segment's correlation structure in other clusters. Higher values indicate that clusters have higher cohesion and better separation, while negative values indicate undesirable clusterings.
For a segment $\segIndex$ with correlation matrix $\correlationMatrix$ the silhouette $\silhouetteIndex$ is defined as
\[\silhouetteIndex = \frac{\intraClusterDistance - \interClusterDistance}{\max(\intraClusterDistance, \interClusterDistance)},\] where $\intraClusterDistance$ is the average distance between all correlation matrices $\correlationMatrix$ in a cluster $C_{\clusterIndex}$ consisting of the subset of segments $\clusterIndices$ \[\intraClusterDistance=\frac{1}{|\clusterIndices| - 1} \sum_{y \in \clusterIndices, y \neq \segIndex} \clusterDistance,\] and $\interClusterDistance$ is the average distance between correlation matrix $\correlationMatrix$ and all other correlation matrices $\correlationMatrix[y]$ in the nearest cluster $C_o$ where $o\neq k$ \[\interClusterDistance=\min \left( \frac{1}{|\clusterIndices[o]|} \sum_{y \in \clusterIndices[o]} \clusterDistance\right).\] From this the SWC is defined as:
\begin{equation}
    \internalIndex[SWC] = \frac{1}{\nSegments}\sum_{\segIndex\in[\nSegments]}\silhouetteIndex,
    \label{eq:scw}
\end{equation}
where $\nSegments$ is the total number of segments and $[\nSegments]$ the set of all the segment indices of a clustering.

\paragraph{DBI - Davies Bouldin Index}
The Davies-Bouldin index (DBI) \cite{Davies1979} is a measure $\geq0$ that evaluates how well different clusters $C_{\clusterIndex}$ are separated from each other and how compact the objects (in our case correlation matrices $\correlationMatrix$) in each cluster are. Lower DBI indicates a compact and better separated clustering. 
For a cluster $C_{\clusterIndex}$, the average distance $\averageClusterDistance$ between the cluster's correlation matrices $\correlationMatrix$, $\segIndex\in\clusterIndices$ (where $\clusterIndices$ is the subset of segments indices in cluster $C_{\clusterIndex}$) and the cluster centroid $\clusterCentroid$ (the correlation matrix of all observations in a cluster) is defined as \[\averageClusterDistance=\frac{1}{|\clusterIndices|}\sum_{m \in \clusterIndices} \clusterDistance[\clusterCentroid].\] From this, the Davies-Bouldin index (DBI) is then defined as:
\begin{equation}
    \internalIndex[DBI] = \frac{1}{\nClusters}\sum_{\clusterIndex\in[\nClusters]}\underset{y\in[\nClusters],y \neq \clusterIndex}{\max}\left(\frac{\averageClusterDistance+\averageClusterDistance[y]}{d(\clusterCentroid, \clusterCentroid[y])}\right),
    \label{eq:dbi}
\end{equation}
where $\nClusters$ is the total number of clusters and $[\nClusters]$ the set of all the cluster indices in a clustering.

\subsubsection{Jaccard Index}
The Jaccard index is an external validity index that is widely used to assess the quality of a clustering $\partition$ created by segmentation or clustering of objects. External validation of a clustering result is only possible when ground-truth clustering $\partition[G]$ is known \cite{Vendramin2010, Arbelaitz2013}. In a real-world setting, this is not the case, and researchers need to rely on internal validation methods.

The Jaccard index is defined as:
\begin{equation}
    J = \frac{\partition[G]\cap \partition}{\partition[G] \cup \partition}
    \label{eq:jaccard-general}
\end{equation}

Equation \ref{eq:jaccard-general} is adjusted to fit the clustering domain and the general definition is \cite{Vendramin2010}:\[ J = \frac{t_p}{t_p+f_n+f_p}\]
When clustering n-dimensional objects $t_p$ is the number of objects that are in the same cluster in clustering $\partition$ and the ground truth clustering $\partition[G]$; $f_n$ is the number of objects belonging to the same cluster in $\partition[G]$ but are in a different cluster in $\partition$; and $f_p$ is the number of objects belonging to different clusters in $\partition[G]$ but are in the same cluster in $\partition$. 
In time series segmentation or change point detection, $t_p$ is the number of change points $\segmentation[\partitionIndex]$ for segmentation $\partition$ that are within a small zone $sz$ around each ground truth change point $\segmentation[G]$ in the ground truth segmentation $\partition[G]$ counting only the first change point in each $sz$; $f_p$ is the number of additional change points $\segmentation[\partitionIndex]$ within a $sz$ and the number of change points in $\partition$ outside of a $sz$; and $f_n$ is the number of $sz$ in segmentation $\partition[G]$ without a change point in $\partition$ \cite{Burg2020,Gensler2014}. For our domain, we are assessing the quality of a clustering $\partition$ as the result of both segmentation and clustering of the segments, we define the Jaccard index as:
\begin{equation}
    J=\frac{t_p}{T}
    \label{eq:jaccard-index}
\end{equation}
where $t_p$ is the number of observations that are in the same cluster in $\partition[G]$ and $\partition$; and $T=t_p+f_p+f_n$ is the total number of observations in the time series. For the simple reason that $f_p+f_n$ is the number of observations that are not in the same cluster in $\partition$ and the ground-truth clustering $\partition[G]$ due to a segmentation or clustering error.

\subsection{Reference Values}
Tables~\ref{tab:benchmark-summary-normal}-\ref{tab:benchmark-summary-downsampled} provide reference values for performance measures under controlled degradation conditions across normal, non-normal, and downsampled data variants, respectively. Each table presents Jaccard index, Silhouette Width Coefficient (SWC), Davies-Bouldin Index (DBI), and MAE measures for specific error conditions, defined by the number of observations shifted or segments assigned to a random wrong cluster. These values serve as calibration points for interpreting the performance of clustering algorithms.

\subsubsection{Correlated (normal distributed) Data Variants}

\begin{table}[!ht]
\caption{Reference table for Jaccard index, SCW, DBI and MAE results achieved for various clustering mistakes (number of observations shifted to the next cluster (obs) and number of segments assigned to a wrong cluster) for the \textbf{normal data variants}.}
\label{tab:benchmark-summary-normal}
\small
\setlength{\tabcolsep}{3pt}
\centering
\begin{tabular}{lrrrrrrrrrrrrrr}
\toprule
\multirow{2}{*}{\textbf{Comp.}} & \multirow{2}{*}{\textbf{obs}} & \multirow{2}{*}{\textbf{clust}} & \multicolumn{3}{c}{\textbf{Jaccard}} & \multicolumn{3}{c}{\textbf{SCW}} & \multicolumn{3}{c}{\textbf{DBI}} & \multicolumn{3}{c}{\textbf{MAE}} \\
\cmidrule(lr){4-6} \cmidrule(lr){7-9} \cmidrule(lr){10-12} \cmidrule(lr){13-15}
&  &  & \textbf{mean} & \textbf{25\%} & \textbf{75\%} & \textbf{mean} & \textbf{25\%} & \textbf{75\%} & \textbf{mean} & \textbf{25\%} & \textbf{75\%} & \textbf{mean} & \textbf{25\%} & \textbf{75\%} \\
\midrule
\multirow{12}{*}{\textbf{100\%}} & 0 & 0 & 1.0 & 1.0 & 1.0 & 0.98 & 0.98 & 0.98 & 0.04 & 0.04 & 0.05 & 0.02 & 0.02 & 0.03 \\
& 50 & 0 & 0.99 & 0.99 & 0.99 & 0.95 & 0.95 & 0.95 & 0.08 & 0.08 & 0.08 & 0.03 & 0.03 & 0.03 \\
& 200 & 0 & 0.98 & 0.98 & 0.98 & 0.80 & 0.80 & 0.80 & 0.28 & 0.28 & 0.28 & 0.06 & 0.06 & 0.06 \\
& 400 & 0 & 0.97 & 0.97 & 0.97 & 0.54 & 0.54 & 0.54 & 0.56 & 0.56 & 0.56 & 0.11 & 0.11 & 0.11 \\
& 0 & 5 & 0.96 & 0.94 & 0.98 & 0.80 & 0.80 & 0.81 & 0.88 & 0.74 & 1.05 & 0.06 & 0.06 & 0.06 \\
& 800 & 0 & 0.94 & 0.94 & 0.94 & 0.23 & 0.22 & 0.25 & 1.13 & 1.08 & 1.20 & 0.19 & 0.18 & 0.20 \\
& 0 & 20 & 0.79 & 0.76 & 0.82 & 0.33 & 0.32 & 0.34 & 2.58 & 2.32 & 2.80 & 0.18 & 0.17 & 0.19 \\
& 0 & 40 & 0.58 & 0.53 & 0.62 & -0.08 & -0.11 & -0.08 & 4.14 & 3.95 & 4.52 & 0.33 & 0.32 & 0.34 \\
& 0 & 60 & 0.40 & 0.37 & 0.42 & -0.28 & -0.30 & -0.26 & 6.32 & 5.54 & 6.57 & 0.46 & 0.45 & 0.47 \\
& 0 & 80 & 0.23 & 0.21 & 0.23 & -0.36 & -0.37 & -0.35 & 7.37 & 6.85 & 7.87 & 0.63 & 0.62 & 0.64 \\
& 800 & 100 & 0.002 & 0.003 & 0.004 & -0.37 & -0.39 & -0.35 & 6.73 & 5.67 & 6.94 & 0.73 & 0.71 & 0.75 \\
& 0 & 100 & 0.0 & 0.0 & 0.0 & -0.38 & -0.41 & -0.36 & 7.05 & 6.33 & 7.62 & 0.77 & 0.76 & 0.79 \\
\midrule
\multirow{10}{*}{\textbf{70\%}} & 0 & 0 & 1.0 & 1.0 & 1.0 & 0.97 & 0.97 & 0.97 & 0.05 & 0.04 & 0.06 & 0.03 & 0.02 & 0.03 \\
& 50 & 0 & 0.99 & 0.99 & 0.99 & 0.94 & 0.94 & 0.94 & 0.10 & 0.10 & 0.10 & 0.03 & 0.03 & 0.03 \\
& 200 & 0 & 0.98 & 0.98 & 0.98 & 0.69 & 0.68 & 0.69 & 0.40 & 0.40 & 0.41 & 0.08 & 0.08 & 0.08 \\
& 0 & 5 & 0.96 & 0.94 & 0.98 & 0.80 & 0.79 & 0.81 & 0.88 & 0.74 & 1.04 & 0.06 & 0.06 & 0.06 \\
& 400 & 0 & 0.96 & 0.96 & 0.96 & 0.42 & 0.42 & 0.42 & 0.82 & 0.82 & 0.82 & 0.13 & 0.13 & 0.13 \\
& 0 & 20 & 0.79 & 0.76 & 0.82 & 0.32 & 0.32 & 0.34 & 2.58 & 2.34 & 2.80 & 0.18 & 0.17 & 0.19 \\
& 0 & 40 & 0.58 & 0.53 & 0.62 & -0.09 & -0.11 & -0.08 & 4.12 & 3.93 & 4.50 & 0.33 & 0.32 & 0.34 \\
& 0 & 60 & 0.40 & 0.37 & 0.42 & -0.28 & -0.31 & -0.26 & 6.31 & 5.58 & 6.58 & 0.46 & 0.45 & 0.47 \\
& 0 & 80 & 0.23 & 0.21 & 0.23 & -0.36 & -0.37 & -0.35 & 7.36 & 6.86 & 8.10 & 0.63 & 0.62 & 0.64 \\
& 0 & 100 & 0.0 & 0.0 & 0.0 & -0.38 & -0.41 & -0.36 & 7.02 & 6.30 & 7.59 & 0.77 & 0.75 & 0.79 \\
\midrule
\multirow{10}{*}{\textbf{10\%}} & 0 & 0 & 1.0 & 1.0 & 1.0 & 0.92 & 0.91 & 0.93 & 0.14 & 0.13 & 0.15 & 0.03 & 0.03 & 0.04 \\
& 50 & 0 & 0.96 & 0.96 & 0.96 & 0.44 & 0.43 & 0.45 & 0.74 & 0.71 & 0.75 & 0.13 & 0.12 & 0.14 \\
& 0 & 5 & 0.96 & 0.94 & 0.98 & 0.75 & 0.73 & 0.76 & 0.93 & 0.77 & 1.09 & 0.07 & 0.07 & 0.07 \\
& 100 & 0 & 0.92 & 0.92 & 0.92 & 0.13 & 0.12 & 0.15 & 1.40 & 1.33 & 1.42 & 0.23 & 0.22 & 0.24 \\
& 0 & 20 & 0.79 & 0.76 & 0.82 & 0.29 & 0.28 & 0.30 & 2.63 & 2.38 & 2.90 & 0.19 & 0.18 & 0.20 \\
& 0 & 40 & 0.58 & 0.53 & 0.62 & -0.10 & -0.12 & -0.08 & 4.16 & 4.08 & 4.63 & 0.34 & 0.33 & 0.35 \\
& 0 & 60 & 0.40 & 0.37 & 0.42 & -0.28 & -0.30 & -0.26 & 6.23 & 5.51 & 6.67 & 0.47 & 0.46 & 0.48 \\
& 0 & 80 & 0.23 & 0.21 & 0.23 & -0.36 & -0.37 & -0.35 & 7.26 & 7.00 & 7.78 & 0.63 & 0.62 & 0.64 \\
& 100 & 100 & 0.004 & 0.002 & 0.005 & -0.37 & -0.39 & -0.35 & 6.55 & 5.57 & 7.0 & 0.73 & 0.72 & 0.76 \\
& 0 & 100 & 0.0 & 0.0 & 0.0 & -0.38 & -0.40 & -0.36 & 7.00 & 6.14 & 7.72 & 0.77 & 0.76 & 0.79 \\
\bottomrule
\end{tabular}
\end{table}

\clearpage

\subsubsection{Non-normal (correlated) Data Variants}

\begin{table}[!ht]
\caption{Reference table for Jaccard index, SCW, DBI and MAE results achieved for various clustering mistakes (number of observations shifted to the next cluster (obs) and number of segments assigned to a wrong cluster) for the \textbf{non-normal data variants}.}
\label{tab:benchmark-summary-non-normal}
\small
\setlength{\tabcolsep}{3pt}
\centering
\begin{tabular}{lrrrrrrrrrrrrrr}
\toprule
\multirow{2}{*}{\textbf{Comp.}} & \multirow{2}{*}{\textbf{obs}} & \multirow{2}{*}{\textbf{clust}} & \multicolumn{3}{c}{\textbf{Jaccard}} & \multicolumn{3}{c}{\textbf{SCW}} & \multicolumn{3}{c}{\textbf{DBI}} & \multicolumn{3}{c}{\textbf{MAE}} \\
\cmidrule(lr){4-6} \cmidrule(lr){7-9} \cmidrule(lr){10-12} \cmidrule(lr){13-15}
&  &  & \textbf{mean} & \textbf{25\%} & \textbf{75\%} & \textbf{mean} & \textbf{25\%} & \textbf{75\%} & \textbf{mean} & \textbf{25\%} & \textbf{75\%} & \textbf{mean} & \textbf{25\%} & \textbf{75\%} \\
\midrule
\multirow{12}{*}{\textbf{100\%}} & 0 & 0 & 1.0 & 1.0 & 1.0 & 0.98 & 0.98 & 0.98 & 0.04 & 0.04 & 0.05 & 0.02 & 0.02 & 0.03 \\
& 50 & 0 & 0.99 & 0.99 & 0.99 & 0.95 & 0.95 & 0.95 & 0.08 & 0.08 & 0.08 & 0.03 & 0.03 & 0.03 \\
& 200 & 0 & 0.98 & 0.98 & 0.98 & 0.80 & 0.80 & 0.80 & 0.28 & 0.28 & 0.28 & 0.06 & 0.06 & 0.06 \\
& 400 & 0 & 0.97 & 0.97 & 0.97 & 0.54 & 0.54 & 0.54 & 0.56 & 0.56 & 0.56 & 0.11 & 0.11 & 0.11 \\
& 0 & 5 & 0.96 & 0.94 & 0.98 & 0.80 & 0.80 & 0.81 & 0.88 & 0.74 & 1.04 & 0.06 & 0.06 & 0.06 \\
& 800 & 0 & 0.94 & 0.94 & 0.94 & 0.23 & 0.22 & 0.25 & 1.13 & 1.08 & 1.20 & 0.19 & 0.18 & 0.20 \\
& 0 & 20 & 0.79 & 0.76 & 0.82 & 0.33 & 0.32 & 0.34 & 2.58 & 2.32 & 2.80 & 0.18 & 0.17 & 0.19 \\
& 0 & 40 & 0.58 & 0.53 & 0.62 & -0.08 & -0.11 & -0.08 & 4.14 & 3.96 & 4.51 & 0.33 & 0.32 & 0.34 \\
& 0 & 60 & 0.40 & 0.37 & 0.42 & -0.28 & -0.30 & -0.26 & 6.36 & 5.55 & 6.62 & 0.46 & 0.45 & 0.47 \\
& 0 & 80 & 0.23 & 0.21 & 0.23 & -0.36 & -0.37 & -0.35 & 7.37 & 6.86 & 7.97 & 0.63 & 0.61 & 0.64 \\
& 800 & 100 & 0.003 & 0.002 & 0.004 & -0.37 & -0.39 & -0.35 & 6.73 & 5.67 & 6.96 & 0.73 & 0.71 & 0.75 \\
& 0 & 100 & 0.0 & 0.0 & 0.0 & -0.38 & -0.41 & -0.36 & 7.05 & 6.33 & 7.61 & 0.77 & 0.75 & 0.78 \\
\midrule
\multirow{10}{*}{\textbf{70\%}} & 0 & 0 & 1.0 & 1.0 & 1.0 & 0.97 & 0.97 & 0.97 & 0.05 & 0.05 & 0.06 & 0.03 & 0.02 & 0.03 \\
& 50 & 0 & 0.99 & 0.99 & 0.99 & 0.94 & 0.94 & 0.94 & 0.10 & 0.10 & 0.10 & 0.03 & 0.03 & 0.03 \\
& 200 & 0 & 0.98 & 0.98 & 0.98 & 0.69 & 0.68 & 0.69 & 0.40 & 0.40 & 0.41 & 0.08 & 0.08 & 0.08 \\
& 0 & 5 & 0.96 & 0.94 & 0.98 & 0.80 & 0.79 & 0.81 & 0.88 & 0.74 & 1.04 & 0.06 & 0.06 & 0.06 \\
& 400 & 0 & 0.96 & 0.96 & 0.96 & 0.42 & 0.42 & 0.42 & 0.82 & 0.82 & 0.82 & 0.13 & 0.13 & 0.13 \\
& 0 & 20 & 0.79 & 0.76 & 0.82 & 0.32 & 0.32 & 0.34 & 2.58 & 2.34 & 2.80 & 0.18 & 0.17 & 0.19 \\
& 0 & 40 & 0.58 & 0.53 & 0.62 & -0.09 & -0.11 & -0.08 & 4.12 & 3.93 & 4.50 & 0.33 & 0.32 & 0.34 \\
& 0 & 60 & 0.40 & 0.37 & 0.42 & -0.28 & -0.31 & -0.26 & 6.33 & 5.59 & 6.64 & 0.46 & 0.45 & 0.47 \\
& 0 & 80 & 0.23 & 0.21 & 0.23 & -0.36 & -0.37 & -0.35 & 7.36 & 6.86 & 8.16 & 0.63 & 0.61 & 0.64 \\
& 0 & 100 & 0.0 & 0.0 & 0.0 & -0.38 & -0.41 & -0.36 & 7.02 & 6.30 & 7.56 & 0.77 & 0.75 & 0.78 \\
\midrule
\multirow{10}{*}{\textbf{10\%}} & 0 & 0 & 1.0 & 1.0 & 1.0 & 0.92 & 0.91 & 0.93 & 0.14 & 0.13 & 0.15 & 0.03 & 0.03 & 0.04 \\
& 50 & 0 & 0.96 & 0.96 & 0.96 & 0.44 & 0.43 & 0.45 & 0.74 & 0.71 & 0.76 & 0.13 & 0.12 & 0.14 \\
& 0 & 5 & 0.96 & 0.94 & 0.98 & 0.75 & 0.73 & 0.76 & 0.93 & 0.77 & 1.09 & 0.07 & 0.07 & 0.07 \\
& 100 & 0 & 0.92 & 0.92 & 0.92 & 0.13 & 0.11 & 0.15 & 1.40 & 1.33 & 1.42 & 0.23 & 0.22 & 0.24 \\
& 0 & 20 & 0.79 & 0.76 & 0.82 & 0.29 & 0.28 & 0.29 & 2.62 & 2.38 & 2.90 & 0.19 & 0.18 & 0.20 \\
& 0 & 40 & 0.58 & 0.53 & 0.62 & -0.10 & -0.12 & -0.08 & 4.16 & 4.08 & 4.63 & 0.34 & 0.33 & 0.35 \\
& 0 & 60 & 0.40 & 0.37 & 0.42 & -0.28 & -0.30 & -0.26 & 6.25 & 5.50 & 6.72 & 0.47 & 0.46 & 0.48 \\
& 0 & 80 & 0.23 & 0.21 & 0.23 & -0.36 & -0.37 & -0.35 & 7.31 & 7.00 & 7.83 & 0.63 & 0.62 & 0.64 \\
& 100 & 100 & 0.004 & 0.002 & 0.005 & -0.37 & -0.39 & -0.35 & 6.55 & 5.57 & 6.99 & 0.73 & 0.71 & 0.76 \\
& 0 & 100 & 0.0 & 0.0 & 0.0 & -0.38 & -0.40 & -0.36 & 7.00 & 6.14 & 7.71 & 0.77 & 0.76 & 0.79 \\
\bottomrule
\end{tabular}
\end{table}

\clearpage

\subsubsection{Downsampled (non-normal distributed) Data Variants}

\begin{table}[!ht]
\caption{Reference table for Jaccard index, SCW, DBI and MAE results achieved for various clustering mistakes (number of observations shifted to the next cluster (obs) and number of segments assigned to a wrong cluster) for the \textbf{downsampled data variants}.}
\label{tab:benchmark-summary-downsampled}
\small
\setlength{\tabcolsep}{3pt}
\centering
\begin{tabular}{lrrrrrrrrrrrrrr}
\toprule
\multirow{2}{*}{\textbf{Comp.}} & \multirow{2}{*}{\textbf{obs}} & \multirow{2}{*}{\textbf{clust}} & \multicolumn{3}{c}{\textbf{Jaccard}} & \multicolumn{3}{c}{\textbf{SCW}} & \multicolumn{3}{c}{\textbf{DBI}} & \multicolumn{3}{c}{\textbf{MAE}} \\
\cmidrule(lr){4-6} \cmidrule(lr){7-9} \cmidrule(lr){10-12} \cmidrule(lr){13-15}
&  &  & \textbf{mean} & \textbf{25\%} & \textbf{75\%} & \textbf{mean} & \textbf{25\%} & \textbf{75\%} & \textbf{mean} & \textbf{25\%} & \textbf{75\%} & \textbf{mean} & \textbf{25\%} & \textbf{75\%} \\
\midrule
\multirow{10}{*}{\textbf{100\%}} & 0 & 0 & 1.0 & 1.0 & 1.0 & 0.63 & 0.58 & 0.67 & 0.50 & 0.45 & 0.56 & 0.13 & 0.12 & 0.15 \\
& 0 & 5 & 0.96 & 0.94 & 0.98 & 0.46 & 0.43 & 0.48 & 1.14 & 1.02 & 1.29 & 0.18 & 0.17 & 0.18 \\
& 50 & 0 & 0.80 & 0.80 & 0.80 & -0.16 & -0.17 & -0.15 & 2.87 & 2.63 & 3.12 & 0.39 & 0.38 & 0.40 \\
& 0 & 20 & 0.79 & 0.76 & 0.82 & 0.09 & 0.06 & 0.12 & 2.86 & 2.61 & 3.01 & 0.27 & 0.26 & 0.28 \\
& 100 & 0 & 0.65 & 0.65 & 0.66 & -0.28 & -0.29 & -0.27 & 4.88 & 4.43 & 5.39 & 0.48 & 0.46 & 0.50 \\
& 0 & 40 & 0.58 & 0.53 & 0.62 & -0.18 & -0.20 & -0.16 & 4.16 & 3.77 & 4.66 & 0.38 & 0.36 & 0.39 \\
& 0 & 60 & 0.40 & 0.37 & 0.42 & -0.31 & -0.33 & -0.29 & 6.26 & 5.41 & 7.21 & 0.48 & 0.47 & 0.48 \\
& 0 & 80 & 0.23 & 0.21 & 0.23 & -0.36 & -0.38 & -0.35 & 7.77 & 6.74 & 8.15 & 0.61 & 0.60 & 0.62 \\
& 100 & 100 & 0.017 & 0.012 & 0.02 & -0.38 & -0.39 & -0.36 & 7.78 & 6.79 & 7.77 & 0.68 & 0.66 & 0.70 \\
& 0 & 100 & 0.0 & 0.0 & 0.0 & -0.38 & -0.40 & -0.36 & 7.31 & 6.43 & 7.61 & 0.72 & 0.70 & 0.74 \\
\midrule
\multirow{10}{*}{\textbf{70\%}} & 0 & 0 & 1.00 & 1.00 & 1.00 & 0.63 & 0.59 & 0.66 & 0.49 & 0.45 & 0.52 & 0.13 & 0.11 & 0.15\\
& 0 & 5 & 0.96 & 0.94 & 0.98 & 0.47 & 0.45 & 0.49 & 1.13 & 1.00 & 1.28 & 0.17 & 0.16 & 0.17\\
& 50 & 0 & 0.80 & 0.80 & 0.80 & -0.16 & -0.18 & -0.14 & 2.98 & 2.72 & 3.24 & 0.39 & 0.38 & 0.40\\
& 0 & 20 & 0.79 & 0.76 & 0.82 & 0.10 & 0.07 & 0.12 & 2.92 & 2.55 & 3.14 & 0.27 & 0.26 & 0.28\\
& 100 & 0 & 0.65 & 0.65 & 0.66 & -0.28 & -0.29 & -0.27 & 5.04 & 4.39 & 4.75 & 0.48 & 0.46 & 0.49\\
& 0 & 40 & 0.58 & 0.53 & 0.62 & -0.17 & -0.19 & -0.16 & 4.18 & 3.93 & 4.71 & 0.37 & 0.36 & 0.39\\
& 0 & 60 & 0.40 & 0.37 & 0.42 & -0.31 & -0.33 & -0.29 & 6.31 & 5.61 & 6.52 & 0.48 & 0.46 & 0.49\\
& 0 & 80 & 0.23 & 0.21 & 0.23 & -0.36 & -0.38 & -0.35 & 7.48 & 6.73 & 8.10 & 0.61 & 0.60 & 0.62\\
& 100 & 100 & 0.02 & 0.01 & 0.02 & -0.38 & -0.40 & -0.35 & 7.45 & 6.55 & 7.85 & 0.68 & 0.66 & 0.70\\
& 0 & 100 & 0.00 & 0.00 & 0.00 & -0.38 & -0.39 & -0.36 & 7.28 & 6.15 & 7.86 & 0.73 & 0.71 & 0.74\\
\midrule
\multirow{10}{*}{\textbf{10\%}} & 0 & 0 & 1.00 & 1.00 & 1.00 & 0.67 & 0.65 & 0.70 & 0.44 & 0.41 & 0.47 & 0.11 & 0.10 & 0.12\\
& 0 & 5 & 0.96 & 0.94 & 0.98 & 0.53 & 0.52 & 0.53 & 1.08 & 0.91 & 1.26 & 0.14 & 0.14 & 0.15\\
& 50 & 0 & 0.80 & 0.80 & 0.80 & -0.15 & -0.17 & -0.14 & 3.08 & 2.67 & 3.13 & 0.39 & 0.38 & 0.40\\
& 0 & 20 & 0.79 & 0.76 & 0.82 & 0.14 & 0.12 & 0.17 & 2.73 & 2.61 & 2.88 & 0.25 & 0.24 & 0.26\\
& 100 & 0 & 0.65 & 0.65 & 0.66 & -0.28 & -0.29 & -0.26 & 4.79 & 4.28 & 4.93 & 0.48 & 0.46 & 0.50\\
& 0 & 40 & 0.58 & 0.53 & 0.62 & -0.16 & -0.18 & -0.15 & 4.23 & 4.00 & 4.53 & 0.37 & 0.36 & 0.37\\
& 0 & 60 & 0.40 & 0.37 & 0.42 & -0.30 & -0.32 & -0.29 & 6.16 & 5.50 & 7.11 & 0.47 & 0.46 & 0.48\\
& 0 & 80 & 0.23 & 0.21 & 0.23 & -0.36 & -0.37 & -0.35 & 7.15 & 6.29 & 7.50 & 0.62 & 0.60 & 0.63\\
& 100 & 100 & 0.02 & 0.01 & 0.02 & -0.38 & -0.39 & -0.35 & 7.66 & 6.77 & 8.02 & 0.70 & 0.67 & 0.71\\
& 0 & 100 & 0.00 & 0.00 & 0.00 & -0.38 & -0.40 & -0.36 & 7.11 & 6.46 & 7.48 & 0.74 & 0.73 & 0.76\\
\bottomrule
\end{tabular}
\end{table}

\clearpage

\section{Case Study: TICC Algorithm Evaluation}\label{appendix:use-case-experiments}

\subsection{TICC Algorithm Description}
We provide a brief explanation of TICC and refer the user to the paper \cite{Hallac2017} that introduced the algorithm for more details. TICC represents each cluster as a Gaussian inverse covariance matrix that forms a Markov Random Field (MRF). These MRFs capture conditional dependencies between variables, where non-zero elements in the precision matrix indicate direct relationships. The algorithm employs an EM-like approach with the following overall optimisation objective:

\begin{equation}
    \underset{\Theta \in \mathcal{T}, \,\text{P}}{\arg\min}\sum_{i=1}^{K}\left[\overbrace{\| \lambda \circ \Theta_i \|_1}^{\text{sparsity}} + \sum_{X_t \in P_i} \left( \overbrace{-\ell\ell(X_t,\Theta_i)}^{\text{log likelihood}} + \overbrace{\beta \mathds{1} \{X_{t-1}\notin P_i\}}^{\text{temporal consistency}}\right)\right]
    \label{eq:ticc-overall-optimisation-objective}
\end{equation}

$K$ are the numbers of clusters, $\mathcal{T}=\{\Theta_1,...,\Theta_K\}$ are the $K$ inverse covariance matrices $\Theta_i \in \mathbb{R}^{nw \times nw}$ for each cluster $i$, $n$ is the number of time series variates, $w\ll T$ is the window size that determines the subsequence, $T$ the length of the time series, and $\lambda$ is a parameter controlling sparsity. The log-likelihood term:
\begin{equation}
    \ell\ell(X_t,\Theta_i)= -\frac{1}{2}(X_t-\mu_i)^T\Theta_i(X_t-\mu_i)+\frac{1}{2}log\,det\,\Theta_i-\frac{n}{2}log(2\pi),
    \label{eq:ticc-loglikelihood}
\end{equation}
measures how well an observation fits a cluster, while the indicator function  $\mathds{1} \{X_{t-1}\notin P_i\}$ penalizes switching between clusters. $X_t\in \mathbb{R}^{nw}$ is a small subsequence $x_{t-w+1}, ..., x_t$, $\mathds{1} \{X_{t-1}\notin P_i\}$ is the indicator function that checks if the previous window was in the same cluster, $P_i$ are the observations assigned to cluster $i$, and $\beta$ is the cluster switch penalty that enforces temporal consistency. In the E-step, the algorithm updates $P_i$, in the M-step it uses sparse inverse covariance estimation (Toeplitz graphical lasso) to learn the MRFs $\Theta_i$.

\subsection{Method and Experimental Setup}
We evaluated TICC following the standardised protocol described in Section~\ref{sec:benchmark-usage}. For all experiments, we used the following fixed hyperparameters: clusters=$23$, window=$5$, switch penalty=$400$, lambda=$0.11$, max iterations=$10$. These parameters are close to the original parameters and tuning them might improve the outcomes of TICC which was not the goal of our case study. The number of iterations was intentionally limited to reduce computational time, although TICC did not converge even with extended runs of $100$ iterations. This non-convergence behaviour is itself an important finding about TICC's performance on our benchmark dataset and warrants future investigation.

\paragraph{1. Selecting Data Variants} We conducted the evaluation across six data variants, including the normal and non-normal distribution types with each three completeness levels (100\%, 70\%, 10\%). This selection allowed us to assess TICC's sensitivity to both distributional assumptions and data sparsity conditions, essential aspects when evaluating time series clustering algorithms. Many other research questions would be possible such as its sensitivity to number of clusters and segment using the reduced data variants, etc.

\paragraph{2. Generating Clustering Results} For each data variant, we trained TICC on the exploratory subject 'unique-puddle-26' to learn the Markov Random Fields (MRFs $\Theta_i$) that in TICC represent each cluster's relationship structure. Once these MRFs were learnt, we applied the trained models to the remaining $29$ subjects without retraining. The application step only involved the assignment step (E-step) of the algorithm, where each observation is assigned to the best-fitting MRF based on log-likelihood and temporal consistency, without modifying the MRFs themselves. This approach rigorously tested TICC's ability to learn generalisable correlation structures and consistently identify the same patterns across multiple independent time series with identical ground truth. We then translated the results into the same structure \dbname{}'s labels are employing, which includes calculating the achieved Spearman correlations for each segment TICC discovered. This will be used for mapping TICC's custers with the ground truth clusters.

\paragraph{3. Calculating Evaluation Measures} For each data variant, we first mapped the clusters discovered by TICC to the ground truth correlation patterns of \dbname{} by grouping all segments assigned to each cluster identified by TICC and calculating the median value for each coefficient position across the correlation vectors of the segments. This preserves the distinct correlation structure of each individual segment. This median-based approach ensures that even if a cluster contains segments with heterogeneous correlation structures, the representative pattern for a cluster is not dominated by the largest segments. We mapped these median correlation vectors to the empirically achieved correlations from the ground truth segmentation (not to theoretical target patterns), using the L1 norm distance. This approach holds algorithms only to standards actually achievable in practice and takes into account structure variations due to sparsification, distribution shifting, and downsampling. Clusters were considered matches if all median correlation coefficients were within a tolerance of $\pm0.1$ of the median mapped ground truth pattern.

We then calculated two sets of evaluation metrics. The clustering quality measures included Silhouette Width Coefficient (SWC) and Davies-Bouldin Index (DBI) using the L5 norm distance, Jaccard Index as an external validation measure, and Mean Absolute Error (MAE) between empirical and ground truth Spearman correlations. Additionally, we computed pattern and segmentation measures: pattern discovery rate (percentage of ground truth structures with at least one matching cluster within $\pm0.1$), pattern specificity (percentage of identified clusters matching exactly one ground truth structure within $\pm0.1$), segmentation ratio (ratio of algorithm-detected segments to ground truth segments), and segment length ratio (ratio of median segment lengths).

\paragraph{4. Interpretation} To interpret and contextualise the results, we used the reference tables for \dbname{} (see  Appendix~\ref{app:benchmark-reference-values}). These tables provide both ground truth baselines and systematically degraded results with known numbers of misclassified observations and misassigned segments for the clustering quality measures for each data variant. 

\paragraph{5. Statistical Validation} Statistical validation was performed using Wilcoxon signed rank tests with a two-sided alternative hypothesis. We tested three hypotheses investigating whether the differences in SWC between the normal and non-normal data variants were significant, as well as whether the differences between the normal complete and partial, respectively, complete and sparse data variants were significant. We had $30$ subjects in each of the 6 data variants. A Bonferroni correction was applied to account for multiple tests, resulting in an alpha level of $\alpha=0.0167$ ($0.05/3$). All statistical tests were conducted using the exact mode for computation. Differences were precalculated and differences smaller than $10^{-8}$ between paired results were not considered in the analysis. For each hypothesis, the effect size was calculated and the sample size required to achieve 80\% power was determined.

\subsection{Results}
Table \ref{tab:evaluation-results-gt-ticc} gives the mean values and ranges for the Silhouette Width Criterion (SWC), Davies-Bouldin index (DBI), Jaccard index, and mean absolute error(MAE) between the target relaxed correlation structure and the achieved correlation structure. Note that the high DBI value for TICC in the non-normal, 10\% completeness data variant is due to a division by a very small number that indicates that the distance between two different cluster centroids was close to $0$.
\begin{table}[!ht]
\caption{Mean and range of evaluation measures for the $30$ subject for the ground truth clustering and untuned TICC clustering results for the \textbf{normal and non-normal data variants}.}
\label{tab:evaluation-results-gt-ticc}
\small
\setlength{\tabcolsep}{4pt}
\centering
\begin{tabular}{llccccc}
\toprule
\textbf{Completeness} & \textbf{Measure} & \multicolumn{2}{c}{\textbf{Normal}} & \multicolumn{2}{c}{\textbf{Non-normal}} \\
\cmidrule(lr){3-4} \cmidrule(lr){5-6}
 & & \textbf{Ground Truth} & \textbf{TICC} & \textbf{Ground Truth} & \textbf{TICC} \\
\midrule
\textbf{100\%} & SWC & 0.97 (0.97-0.98) & 0.73 (0.56-0.76) & 0.97 (0.97-0.98) & -0.15 (-0.35-0.02) \\
 & DBI & 0.04 (0.03-0.05) & 0.74 (0.67-0.81) & 0.04 (0.03-0.05) & 2.85 (1.35-5.08) \\
 & Jaccard & 1 & 0.82 (0.76-0.89) & 1 & 0.38 (0.09-0.76) \\
 & MAE& 0.02 (0.02-0.03) & 0.07 (0.06-0.07) & 0.02 (0.02-0.03) & 0.15 (0.11-0.19) \\
\midrule
\textbf{70\%} & SWC & 0.97 (0.96-0.98) & 0.72 (0.49-0.76) & 0.97 (0.97-0.98) & -0.15 (-0.35-0.02) \\
 & DBI & 0.05 (0.04-0.07) & 0.54 (0.48-0.79) & 0.04 (0.03-0.05) & 3.59 (2.01-17.02) \\
 & Jaccard & 1 & 0.79 (0.72-0.85) & 1 & 0.38 (0.08-0.72) \\
 & MAE& 0.02 (0.02-0.03) & 0.05 (0.05-0.06) & 0.02 (0.01-0.03) & 0.15 (0.1-0.19) \\
\midrule
\textbf{10\%} & SWC & 0.92 (0.89-0.94) & 0.61 (0.41-0.66) & 0.92 (0.89-0.94) & -0.08 (-0.37-0.16) \\
 & DBI & 0.14 (0.09-0.18) & 1.04 (0.41-0.66) & 0.14 (0.1-0.19) & $>$19Mio, min 1.17 \\
 & Jaccard & 1 & 0.80 (0.7-0.84) & 1 & 0.28 (0.09-0.52) \\
 & MAE& 0.02(0.02-0.03) & 0.07 (0.06-0.08) & 0.03 (0.02-0.04) & 0.14 (0.09-0.23) \\
\bottomrule
\end{tabular}
\end{table}

Table \ref{tab:ticc-pattern-discovery-results} shows the pattern discovery and segmentation performance of TICC. From these results, the TICC segment length and segmentation ratio compare well with the ground truth for the normal data. For non-normal data, TICC over segments.

\begin{table}[!ht]
\caption{Mean and ranges for pattern discovery and segmentation performance of TICC for the $30$ subjects.}
\label{tab:ticc-pattern-discovery-results}
\small
\setlength{\tabcolsep}{4pt}
\centering
\begin{tabular}{llrrrrr}
\toprule
\textbf{Completeness} & \textbf{Measure} & \multicolumn{2}{c}{\textbf{Normal}} & \multicolumn{2}{c}{\textbf{Non-normal}} \\
\cmidrule(lr){3-4} \cmidrule(lr){5-6}
 & & \textbf{mean} & \textbf{range} & \textbf{mean} & \textbf{range} \\
\midrule
\textbf{100\%} & Pattern discovery \% & 81.2 & 78.3-91.3 & 49.6 & 21.7-739 \\
 & Pattern specificity \% & 91 & 56.5-100 & 54.2 & 25-75 \\
 & Segmentation ratio & 0.99 & 0.96-1.07 & 22.7 & 1.2-100 \\
 & Segment length ratio & 0.99 & 0.67-1 & 0.1 & 0-0.5 \\
 & n Cluster & 19.3 & 19-23 & 15.6 & 6-23 \\
\midrule
\textbf{70\%} & Pattern discovery \% & 78.4 & 78.3-82.6 & 50.3 & 17.4-78.3 \\
 & Pattern specificity \% & 98.4 & 56.5-100 & 56.8 & 28.6-100 \\
 & Segmentation ratio & 1 & 0.9-1.1 & 16 & 1.3-74.3 \\
 & Segment length ratio & 1 & 0.7-1 & 0.1 & 0-0.3 \\
 & n Cluster & 18.3 & 18-23 & 15.1 & 6-23 \\
\midrule
\textbf{10\%} & Pattern discovery \% & 78.4 & 78.3-82.6 & 37.5 & 17.4-65.2 \\
 & Pattern specificity \% & 87.4 & 65.2-100 & 63.2 & 16.7-100 \\
 & Segmentation ratio & 1 & 0.9-1 & 2.7 & 0.5-11.7 \\
 & Segment length ratio & 1 & 1-1.2 & 0.4 & 0.1-1.3 \\
 & n Cluster & 19.3 & 18-23 & 10.8 & 4-23 \\
\bottomrule
\end{tabular}
\end{table}

For the normal data variants, TICC consistently struggled to detect certain correlation structures. Structures with higher MAE values (e.g. patterns 7, 15, 24, 4, 5, 8) are harder for TICC to identify. There are notable contradictions such as pattern 18 $[-1, 0, 0]$ and pattern 23 $[-1, 1, -1]$, which have low MAE values, but were consistently missed by TICC. This indicates that while TICC's pattern detection performance is partly influenced by data quality, the algorithm also struggles with certain correlation structures despite their accurate representation in the underlying data. TICC finds it easier to identify simple correlation structures, particularly pattern 0 $[0,0,0]$ that is always detected across all data variants, along with pattern 13 $[1,1,1]$ and patterns with single strong correlations such as patterns 3 $[0,1,0]$ and 9 $[1,0,0]$. TICC performs better for complete data, where it sometimes misses only 2-3 patterns. As data sparsity increases, the consistency of pattern detection decreases.

\begin{table}[!ht]
    \caption{Statistical comparison of SWC performance across different data variants using Wilcoxon signed rank tests with Bonferroni corrected $\alpha=0.0167$.}
    \label{tab:swc-statistical-analysis}
    \centering
  \begin{tabular}{l@{\hspace{6pt}}r@{\hspace{6pt}}r@{\hspace{6pt}}r@{\hspace{6pt}}r}
    \toprule
    \textbf{Hypothesis} & \textbf{p-value} & \textbf{Effect Size} & \textbf{Power (\%)} & \textbf{N for 80\% Power} \\
    \midrule
    H1: SWC normal > non-normal & <0.0001 & 1.146 & >99.99 & 11.0 \\
    H2: SWC complete > partial (normal) & 0.0005 & 0.646 & 81.7 & 28.0 \\
    H3: SWC complete > sparse (normal) & <0.0001 & 1.097 & >99.99 & 12.0 \\
    \bottomrule
  \end{tabular}
\end{table}

Statistical validation results are shown in Table~\ref{tab:swc-statistical-analysis}. For H1 we investigated whether the difference between the normal and non-normal data variants was significant. The results show that SWC performed significantly better with normally distributed data compared to non-normal data ($p < 0.0001$), with a large effect size ($1.146$) and a high statistical power ($>99.99\%$). For H2 we investigated whether the difference between the normal complete and partial data variants was significant. The results were statistically significant ($p = 0.0005$) with a moderate to large effect size ($0.646$). The analysis achieved a power of $81.7\%$ which would require $28$ subjects for this effect size (we have $30$). Finally, for H3 we investigated whether the difference between the normal complete and sparse data variants was significant. The results were also statistically significant ($p < 0.0001$), showing a large effect size ($1.097$), and high statistical power ($>99.99\%$). Overall, these results strongly support the fact that TICC performs best on complete normal data compared to the other 5 data variants. Power and effect size analysis demonstrate that the $30$ subjects for each data variant in \dbname{} provide high power for statistical analyses.
\end{document}